%% file: main.tex
\documentclass[10pt,twocolumn,letterpaper]{article}

\usepackage{iccv}
\usepackage{times}
\usepackage{epsfig}
\usepackage{graphicx}
\usepackage{amsmath}
\usepackage{amssymb}
\usepackage[accsupp]{axessibility} 
\usepackage{xcolor}
\usepackage{multirow}
\usepackage[normalem]{ulem}
\useunder{\uline}{\ul}{}

\usepackage[breaklinks=true,bookmarks=false,colorlinks]{hyperref}
\iccvfinalcopy

\def\iccvPaperID{11074} 

\def\network{V-FUSE\xspace}
\newenvironment{packed_item}{
\begin{itemize}
\vspace{-4pt}
  \setlength{\itemsep}{0pt}
  \setlength{\parskip}{0pt}
  \setlength{\parsep}{0pt}
  \setlength{\topsep}{-10pt}
  \setlength{\partopsep}{0pt}
}{\end{itemize}}

\begin{document}

\title{V-FUSE: Volumetric Depth Map Fusion with Long-Range Constraints}

\author{Nathaniel Burgdorfer \qquad Philippos Mordohai \\ Stevens Institute of Technology}

\maketitle

\input{abstract}

\section{Introduction}
\label{sec:intro}
\input{intro}

\section{Related Work}
\label{sec:related}
\input{related}

\section{Method}
\label{sec:method}
\input{method}

\section{Loss Function}
\label{sec:loss}
\input{loss}

\section{Experiments}
\label{sec:experiments}
\input{experiments}

\section{Conclusion}
\label{sec:conclusion}
\input{conclusion}

\noindent
\textbf{Acknowledgment} This research has been supported in
part by the National Science Foundation under award
2024653.

{\small
\bibliographystyle{ieee_fullname}
\bibliography{egbib}
}

\include{supplement}

\end{document}

%% file: abstract.tex
\begin{abstract}
   We introduce a learning-based depth map fusion framework that accepts a set of depth and confidence maps generated by a Multi-View Stereo (MVS) algorithm as input and improves them. This is accomplished by integrating volumetric visibility constraints that encode long-range surface relationships across different views into an end-to-end trainable architecture. We also introduce a depth search window estimation sub-network trained jointly with the larger fusion sub-network to reduce the depth hypothesis search space along each ray. Our method learns to model depth consensus and violations of visibility constraints directly from the data; effectively removing the necessity of fine-tuning fusion parameters. Extensive experiments on MVS datasets show substantial improvements in the accuracy of the output fused depth and confidence maps. Our code is available at \url{https://github.com/nburgdorfer/V-FUSE}
\end{abstract}

%% file: intro.tex
Much like other areas of computer vision, Multi-View Stereo (MVS) has benefited from the advent of deep learning. Progress has been driven by the creation of end-to-end systems, unifying all aspects of the MVS pipeline, and by replacing heuristics in the components of the pipeline with optimized network modules. An aspect of MVS that requires further investigation is depth map fusion, which is still implemented as a sequence of heuristic operations.

Considering that the top performing MVS systems in terms of geometric accuracy\footnote{NeRF \cite{mildenhall2020nerf} has inspired a vastly expanding class of algorithms that produce superior results in view synthesis, but not in 3D reconstruction. We consider NeRF a separate line of work from MVS.} use depth map collections as the representation, depth map fusion can be a crucial step for obtaining the final 3D reconstruction of the scene. As has been shown by conventional fusion research \cite{Merrell_2007_ICCV}, fusing depth maps, guided by geometric constraints, improves the precision of correct depth estimates by blending them with supporting estimates for the same part of the surface, detects and removes outliers, and reduces redundancy in the final 3D model. Current deep MVS approaches \cite{Cheng_2020_CVPR, Gu_2020_CVPR, Ma_2021_ICCV, Mi_2022_CVPR, Yao_2018_ECCV, Yang_2022_CVPR, Yang_2020_CVPR}, however, bypass depth map fusion and proceed directly to \textit{filtering fusion}, which includes various heuristic post-processing steps to obtain a global point cloud by filtering the point cloud reconstructed from the set of depth maps. This approach has been successful; however, without depth map fusion, not all geometric information from the scene is utilized. Our motivation in this work is to build an end-to-end fusion network that can generate much more accurate depth and confidence maps.

\begin{figure}[t]
\centering
\footnotesize
    \begin{tabular}{cc}
        \includegraphics[width=0.45\columnwidth]{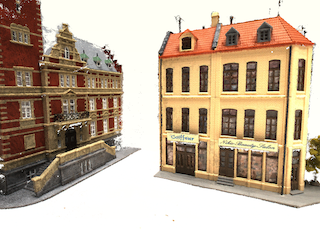} &
        \includegraphics[width=0.45\columnwidth]{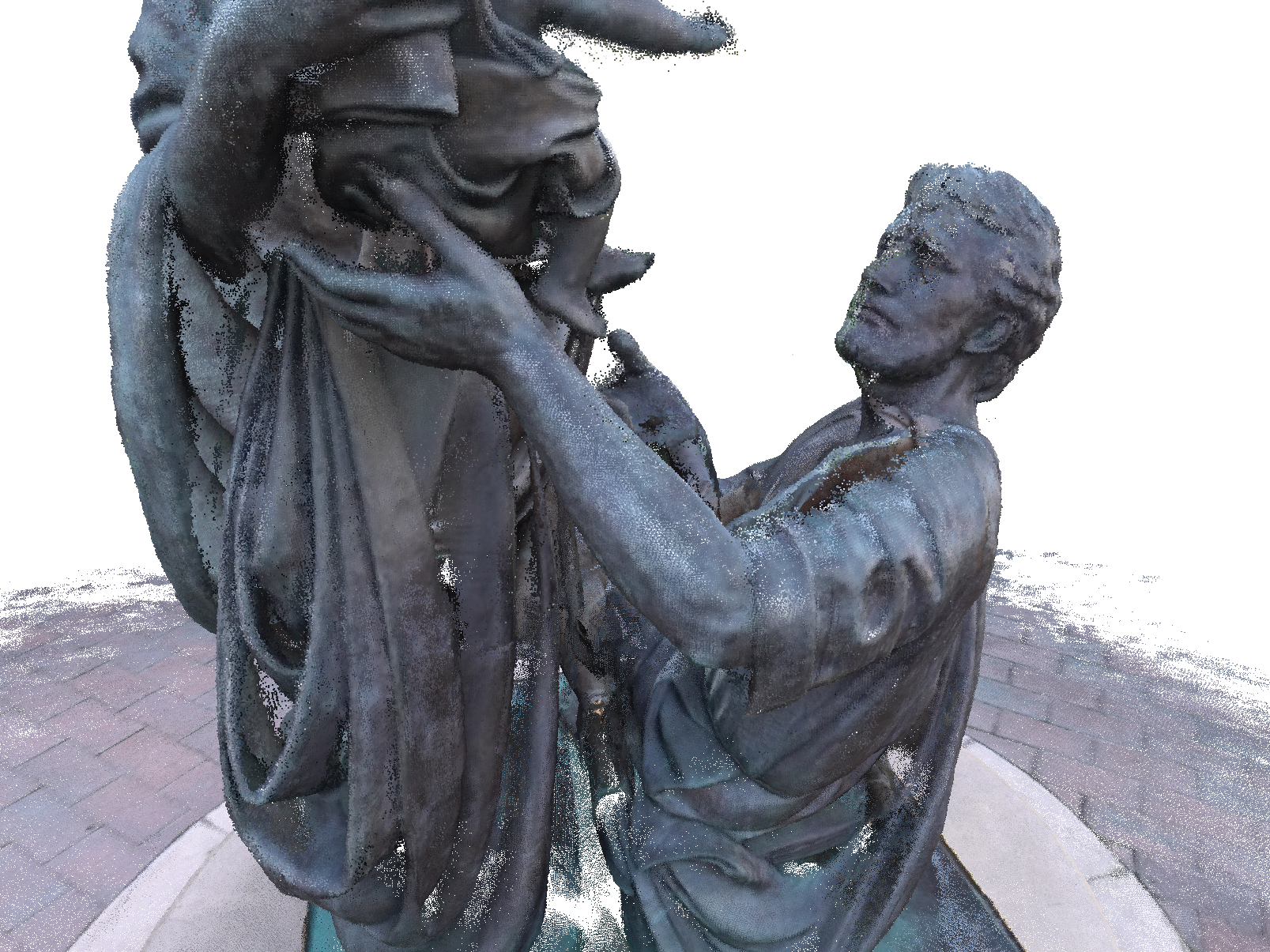} \\
        \includegraphics[width=0.45\columnwidth]{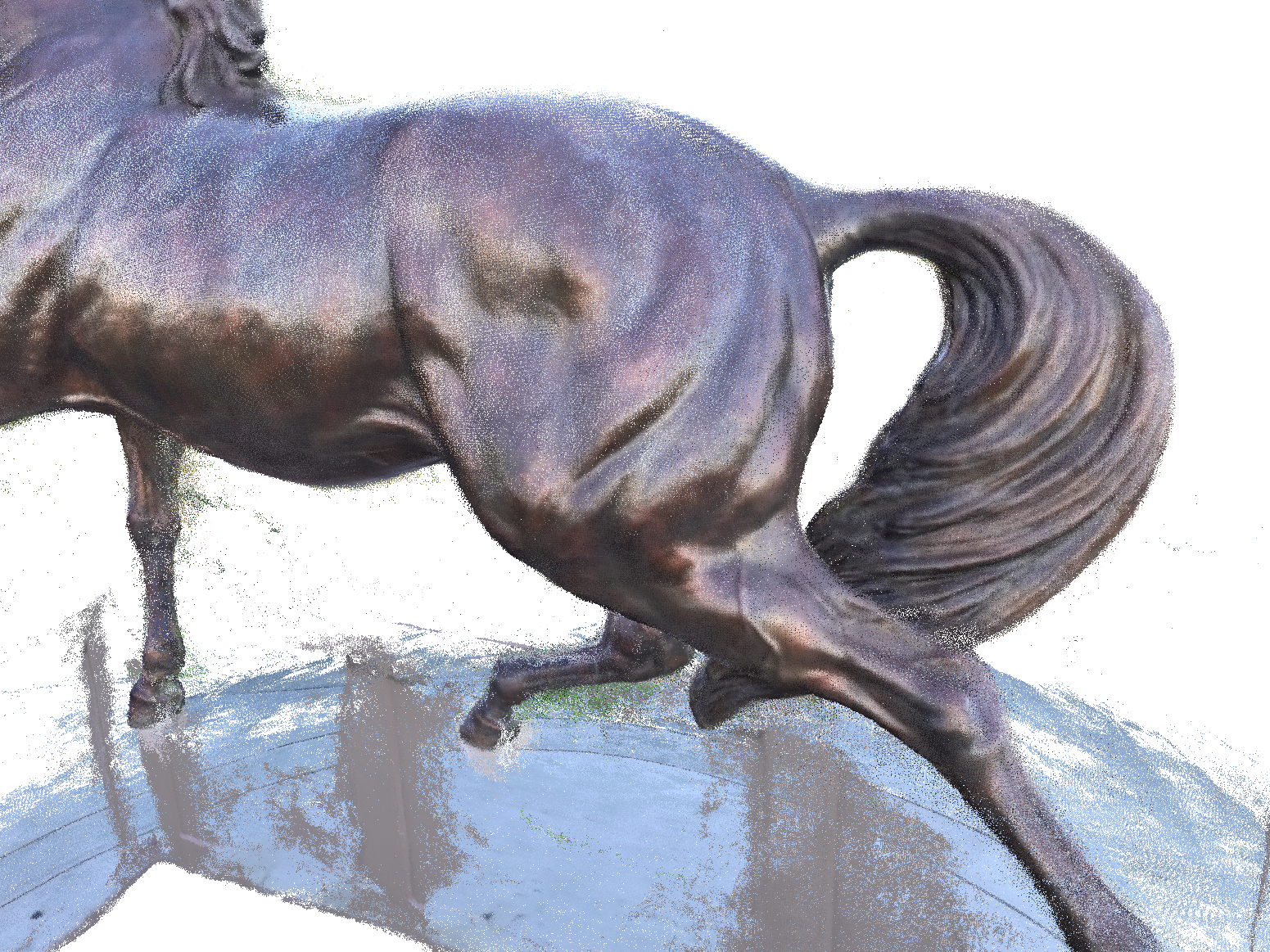} &
        \includegraphics[width=0.45\columnwidth]{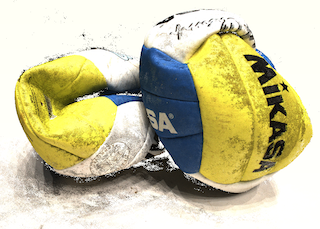} \\
    \end{tabular}
    \caption{Point cloud reconstructions from DTU \cite{aanaes16} and Tanks \& Temples \cite{Knapitsch_2017_TNT} datasets using depth maps from NP-CVP-MVSNet \cite{Yang_2022_CVPR} and UCSNet \cite{Cheng_2020_CVPR} as input to \network{}.}
    \label{fig:scan114}
    \vspace{-10pt}
\end{figure}

Filtering fusion that operates on local 3D neighborhoods is unable to leverage relationships among distant surface primitives, such as a surface being occluded from a faraway object. Similarly, convolution networks have a limited receptive fields and can only reason about local interactions. We present \network{}, an approach that allows a 3D convolutional network to benefit from such geometric information, in a differentiable manner, controlled by learnable hyper-parameters.

\network{} considers three types of constraints, inspired by the work of Merrell et al. \cite{Merrell_2007_ICCV}: \textit{support} among consistent depth estimates across multiple views, \textit{occlusions} and \textit{free-space violations} that provide evidence against depth estimates contradicting surfaces estimated in different depth maps. Free-space violations provide the added benefit of encoding conflicts with respect to surfaces that may be invisible in the frame of the reference camera. There are three substantial differences between our approach and that of Merrell et al.: (i) theirs operates in $2\frac{1}{2}$D while ours operates in a 3D volume, (ii) their algorithms make decisions per pixel without considering context, and (iii) all parameters in our approach are learned end-to-end. Specifying visibility constraints in the fusion volume allows \network{} to reason based on interactions among depth estimates along the rays, as well as spatially among neighboring voxels. In the absence of these constraints, only the latter would have been possible via 3D convolutions, which cannot reason about long-range conflicts.

Reducing the storage and computational requirements of deep MVS networks is a necessity for increasing the resolution and quality of 3D reconstruction. 3D convolutional networks operating on cost volumes are forced to downsample high resolution inputs. Since our framework is also volumetric, we propose a technique for achieving high resolution near the surfaces while keeping memory requirements manageable. Specifically, we learn to generate a per-pixel, narrow depth search window by examining the input depth and confidence estimates. Unlike previous networks that iteratively refine the depth search space, our framework leverages the availability of input depth and confidence estimates to determine a reduced search space in a single pass.
 
Our main contributions are:
\begin{packed_item}
    \item
    An end-to-end learning-based method for the fusion of depth and confidence maps, leveraging long-range, volumetric visibility constraints encoded into a \textit{visibility constraint volume (VCV)}.
    \item
    A pixel-wise search window estimation sub-network to refine the depth  search space.
\end{packed_item}

We provide extensive evaluation of \network{} on MVS benchmarks \cite{aanaes16, Knapitsch_2017_TNT,yao2020blendedmvs}, using 2D and 3D error metrics.

\begin{figure*}[t]
\footnotesize
    \centering
    \begin{tabular}{c}
        \includegraphics[width=0.88\textwidth]{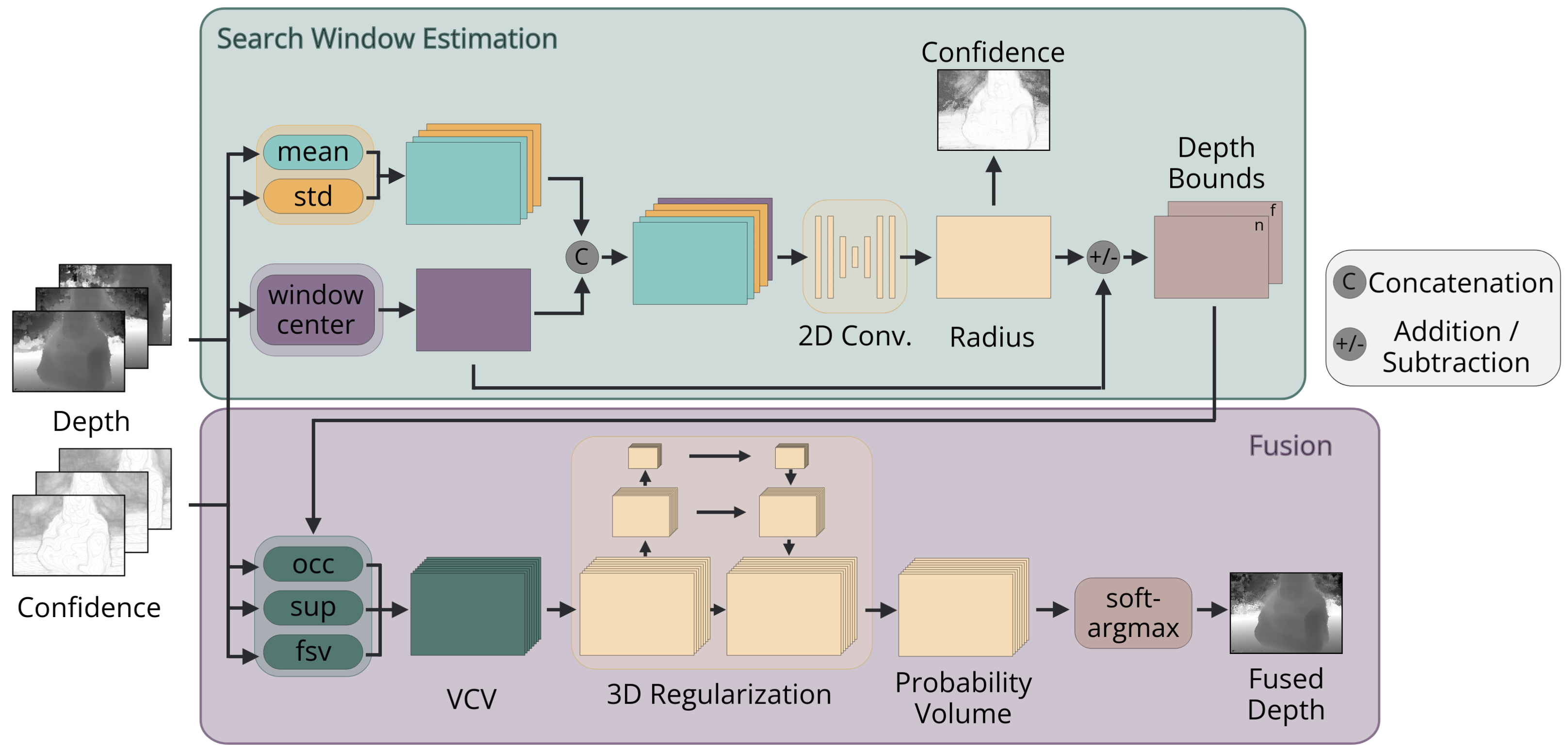}
    \end{tabular}
    \vspace{-2pt}
    \caption{Overview of the \network{} architecture. The network is split into two major sub-networks: \textit{Search Window Estimation (SWE)} and \textit{Fusion}. Both sub-networks take in a set of depth maps and confidence maps for a given set of camera views. The SWE sub-network is responsible for estimating a refined depth search window on each ray of the reference camera. The Fusion sub-network uses these refined depth hypotheses to build a \textit{visibility constraint volume (VCV)}, encoding long-range, volumetric visibility constraints into each voxel of the VCV. After passing this volume through a convolutional network and a soft-argmax operator, we regress the final fused depth map.}
    \label{fig:pipeline}
    \vspace{-10pt}
\end{figure*}

%% file: related.tex
In this section, we review related work on learning-based MVS, as well as conventional and learning-based depth map fusion.(Unfortunately, no recent surveys on these topics are available, to the best of our knowledge.)

The combination of deep learning and plane-sweeping stereo has inspired a new generation of MVS algorithms. The plane-sweeping volume (PSV) \cite{gallup07} allows the use of cost aggregation and disparity estimation techniques developed for binocular stereo \cite{kendall2017gcnet} in multi-view settings. The first deep learning-based plane-sweeping algorithm was DeepStereo \cite{flynn16} that addresses view synthesis in a self-supervised manner. Supervised formulations targeting depth map estimation are largely influenced by MVSNet \cite{Yao_2018_ECCV} and concurrent work \cite{huang2018deepmvs,im2019dpsnet}. We also adopt the PSV structure in this work for our fusion volume.

Several methods \cite{Cheng_2020_CVPR, Gu_2020_CVPR, Ma_2021_ICCV, Mi_2022_CVPR, Wang_2021_CVPR, Yang_2022_CVPR, Yang_2020_CVPR, zhang_2020_BMVC} aim to improve memory efficiency in deep MVS through multi-resolution, iterative schemes that refine the depth search space with each increase in resolution. This is achieved via regular incremental reductions in search range \cite{ Gu_2020_CVPR, Yang_2020_CVPR}, or with a range set using only confidence estimates \cite{Cheng_2020_CVPR}. We have developed a non-iterative method for estimating per-pixel depth search windows based on information extracted from the distribution of input depth and confidence maps.

Recent work has addressed MVS by: combined classification and regression for depth estimation \cite{peng2022rethinking, wang2022itermvs}, sequential depth interval selection \cite{Sormann_2021_BMVC}, an adaptation of RAFT (Recurrent All-Pairs Field Transforms) \cite{ma2022multiview}, operating over adaptive intervals along epipolar lines instead of discrete depths \cite{Ma_2021_ICCV}, and the use of a non-parametric depth distribution model to mitigate shortcomings of unimodal depth models \cite{Yang_2022_CVPR}. Transformers for MVS \cite{ding2022transmvsnet,guizilini2022multi,wang2021multi,wang2022mvster} leverage the intra- and inter-attention mechanisms to achieve more accurate feature matching than previous architectures.

\noindent
\textbf{Conventional Depth Map Fusion}
Conventional fusion methods reduce errors and inconsistencies in MVS pipelines. Merrell et al. \cite{Merrell_2007_ICCV} propose two algorithms for fusing depth maps by selecting depth estimates with large degrees of support from the input depth maps that outweigh violations of visibility constraints. We employ similar constraints, but in a volumetric formulation while the conventional approach \cite{Merrell_2007_ICCV} reasons on $2\frac{1}{2}$D depth maps. Hu and Mordohai \cite{hu2012least} extend the aforementioned method \cite{Merrell_2007_ICCV} by modeling geometric uncertainty, in addition to confidence, and by considering multiple depth candidates per pixel.

A popular choice for fusing depth maps among deep MVS pipelines is the work of Galliani et al. \cite{Galliani_2015_ICCV}. It is based on the projection of depth estimates onto several supporting depth maps to accumulate consensus subject to criteria on reprojection error and surface normal inconsistency. The dense COLMAP pipeline \cite{schonberger2016eccv} also includes a fusion module that rejects outliers based on lack of photometric and geometric support and clusters inliers. Both techniques require setting several thresholds and parameters, and are limited to filtering depth maps into a final 3D model without improving the underlying depth maps.

Some deep MVS systems introduce custom fusion and filtering steps which are not included in the end-to-end trainable pipeline. These include P-MVSNet \cite{Luo_2019_ICCV} that considers pixel and depth reprojection errors, and D$^2$HC-RMVSNet \cite{Yan_2020_ECCV} that includes geometric consistency scores. Instead of relying on filtering and averaging depth estimates, our work aims to refine and fuse depth maps before they are filtered and projected into a point cloud.

\noindent
\textbf{Learning-Based Depth Map Fusion}
Most learning-based fusion methods follow the seminal work of Curless and Levoy \cite{Curless_1996_SIGGRAPH} and adopt a volumetric representation of the truncated signed distance function (TSDF). Learning-based approaches relying on implicit representations \cite{Chen_2019_CVPR,mescheder2019occupancy,park2019deepsdf,peng2020convolutional} model surfaces as continuous decision boundaries of a deep classifier, and are thus ill-suited for open scenes like the ones we reconstruct.

Recent methods \cite{Choe_2021_ICCV, Weder_2020_CVPR, Weder_2021_CVPR} propose fusing streams of input depth maps by learning TSDF volume updates \cite{Weder_2020_CVPR}, by fusing in the latent space and learning a translator to produce a final TSDF volume \cite{Weder_2021_CVPR}, or by learning pose invariant scene volumes jointly with a MVS sub-network \cite{Choe_2021_ICCV}. Volumetric methods suffer from large storage requirements. Our method is volumetric but allows for a very thin volume in the direction of the optical axis.

Donn\'e and Geiger \cite{Donne_2019_CVPR} developed a non-volumetric data-driven approach for fusing depth maps, estimated conventionally or by a learning-based technique. DeFuSR filters out wrong depth estimates, but also improves correct ones via refinement and inpainting sub-networks. It operates on re-projected depth estimates and image features at high resolution, depending entirely on 3D convolutions to reason about consensus.

%% file: method.tex
In this section, we introduce the architecture of \network{} (an overview can be found in Figure \ref{fig:pipeline}). Our network takes as input a reference depth map $D_0 \in \mathbb{R}^{HxW}$ and the corresponding confidence map $C_0 \in \mathbb{R}^{HxW}$, and $N-1$ source depth and confidence maps $\{D_v\}_{v=1}^{N-1} \in \mathbb{R}^{HxW}$ and $\{C_v\}_{v=1}^{N-1} \in \mathbb{R}^{HxW}$ respectively. We begin by rendering the input source maps into the reference view to obtain $\{D^{ref}_v\}_{v=1}^{N-1}$ and $\{C^{ref}_v\}_{v=1}^{N-1}$. With this set of rendered maps, we build a \textit{visibility constraint volume (VCV)}, whose structure follows that of the plane-sweeping volume. The VCV encodes long-range, volumetric, constraints at each voxel. We use a 3D convolutional network to regularize the VCV and regress the final fused depth map. As output, our network produces fused depth and confidence maps for the reference view, $D^f$ and $C^f$. The construction of the VCV and the 3D convolutional network are supervised using an $l_1$ loss between the estimated and ground truth depth maps. For memory and run-time efficiency, we introduce a novel \textit{search window estimation (SWE)} sub-network in order to restrict the depth search space used as input in the construction of the VCV. This sub-network is supervised through a novel loss that we discuss in Section \ref{sec:loss_swe}.

\subsection{Visibility Constraint Volume}
\label{sec:method_vcv}
\begin{figure}[t]
\footnotesize
    \centering
    \begin{tabular}{c}
        \includegraphics[width=0.80\columnwidth]{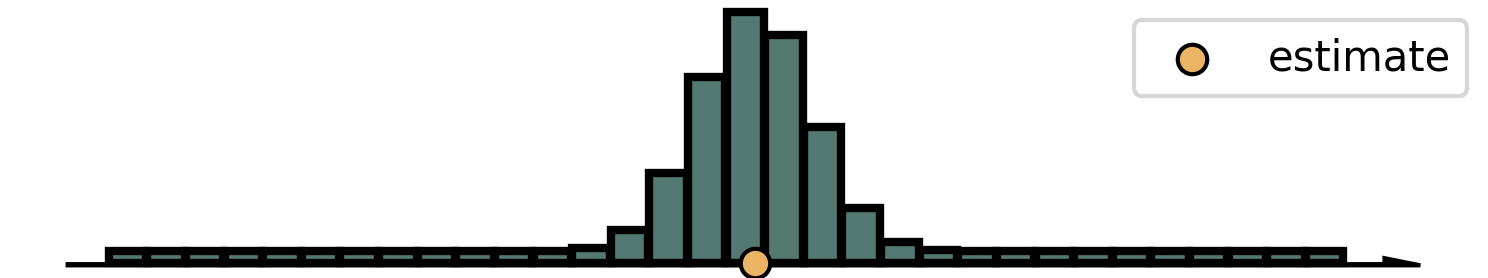} \\
        (a) Support \\
        \includegraphics[width=0.80\columnwidth]{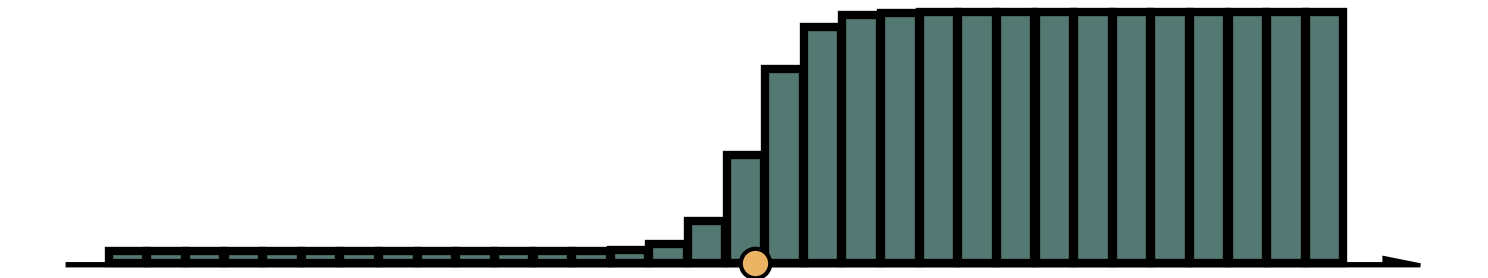} \\
        (b) Occlusions \\
        \includegraphics[width=0.80\columnwidth]{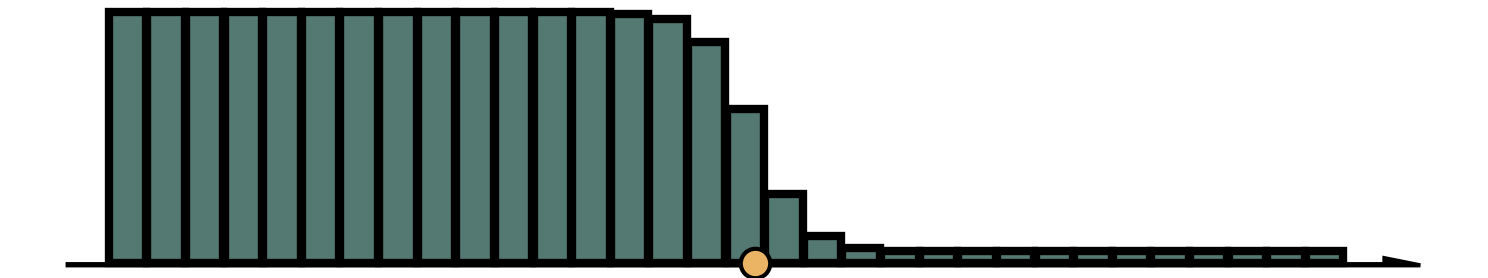} \\
        (c) Free-Space
    \end{tabular}
    \caption{Visualization of each constraint response curve. Support and occlusions are encoded along the ray of the reference camera, while free-space violations are encoded along the ray of each source camera. The constraint curves are centered around each depth estimate. The support response activates near the depth estimate (encoded as a Gaussian). The occlusion response activates past the depth estimate and the free-space response activates before the depth estimate (both encoded as sigmoids).}
    \label{fig:constraints}
\end{figure}

Similar to most deep MVS frameworks, a core component of our network is the construction of a cost volume along the reference camera frustum. However, instead of encoding warped image features, our volume encodes visibility constraints for the purpose of measuring multi-view depth estimate consensus and inconsistency. Specifically, we compute three separate metrics measuring support, occlusions, and free-space violations, and aggregate each metric into separate channels in the VCV. Essentially, each voxel is a collection of the response values for all three constraints from each input view. The constraints are aggregated over all views, each view contributing equally (without favoring the reference view) up to a confidence weighting. We discuss each constraint in detail below. Figure \ref{fig:constraints} shows a visualization of the three constraints.

The network takes as input a set of $M$ initial depth hypotheses $h \in \mathbb{R}^{M}$. This is the set of depth values measured along the ray of the reference camera. We define $p$ to be a given pixel index and $q$ to be the corresponding voxel index at the $d^{th}$ hypothesis. Using the set $h$, we build a hypothesis volume $S \in \mathbb{R}^{HxWxM}$ by tiling $h$ at every pixel, meaning that each ray uses the same set of depth hypotheses. Using $S$, we can compute a 3-channel VCV, $V \in \mathbb{R}^{HxWxMx3}$.

\noindent
\textbf{Support} We first compute the support response for each voxel in the VCV. Intuitively, support is an encoding of the multi-view depth consensus for the rendered depth maps. For a given voxel, the higher the support response, the more probable the true depth value exists at that voxel. For support, we employ a Gaussian distribution centered at each depth estimate in the rendered depth map $D^{ref}_v$ for each view, encoded along the ray of the reference view.

\begin{equation}
    V_{q,0} = \frac{1}{K} \sum_{v=0}^{K-1} C^{ref}_{v,p} \exp \left( { \frac{-(S_q - D^{ref}_{v,p})^2}{2\sigma_p^2} }\right)
\end{equation}
Here, $V_{q,0}$ is the support response for the depth hypothesis at voxel $S_q$. The subscript $0$ is used to indicate that the support response is encoded in the first channel of $V_q$. The confidence $C^{ref}_{v,p}$ rendered into the reference view for view $v$ and pixel $p$ is used to weigh the support response. The standard deviation for the Gaussian distribution $\sigma_p$ is a function of the per-pixel window radius (discussed in Section \ref{sec:method_swe}), which allows the level of support to vary depending on the size of the search window. The formulation of $\sigma_p$ (see supplement) includes a learned hyper-parameter in order for the support window to be learned from the data. Lower values of $\sigma_p$ correspond to a sharper response boundary.

Note that due to perspective distortion and due to some 3D points projected from the source depth maps falling out of bounds, a pixel of the reference view may receive fewer than $N$ rendered depths and confidences. Therefore, we define $K \leq N$ to be the number of views that provide a response for the depth hypothesis at voxel $S_q$. 

\noindent
\textbf{Occlusions} To identify conflicting depth estimates, we include occlusion and free-space violation responses as separate channels in our VCV. Occlusions are events in which the reference depth hypothesis at voxel $S_q$ is farther away from the reference camera than the rendered supporting depth estimate $D^{ref}_{v,p}$. To encode occlusions, we use a sigmoid computed along the ray of the reference view. The sigmoid is centered at each depth estimate in the rendered depth map $D^{ref}_v$ for each view and activates behind the estimate. In this way, the response for occlusions contributes a sigmoid response to $V_q$ with magnitude depending on the difference in depth. The response is high for depth hypotheses that are beyond each estimate and low for hypotheses that are in front of each estimates.

\begin{equation}
    V_{q,1} = \frac{1}{K} \sum_{v=0}^{K-1} C^{ref}_{v,p} \frac{1}{1 + \exp \left( {-\lambda_p (S_q - D^{ref}_{v,p})} \right)}
\end{equation}
Here, $V_{q,1}$ is the occlusion response for the depth hypothesis at voxel $S_q$ encoded into the second channel of $V_q$. The input confidence values are used to weigh the occlusion responses. The multiplier $\lambda_p$ is a function of the per-pixel window radius and is used to adjust the slope of the sigmoid function. The definition for $\lambda_p$ (see supplement) also includes a learned hyper-parameter so that the slope of the sigmoid is learned from the data.

\noindent
\textbf{Free-Space Violations}
In contrast to support and occlusion, free-space violations are measured with respect to the source views. They occur when a depth hypothesis $S^v_q$ (rendered into the source view $v$) is closer to the source camera than the depth estimate from the original, non-rendered source depth map $D_{v,p}$. In this context, we state that the depth hypothesis $S^v_q$ violates the free-space of depth estimate $D_{v,p}$. Much like occlusions, we use a sigmoid to encode free-space violations. The sigmoid function is defined along the ray of projection of the original depth map $D_v$ for each view and activates in front of the depth estimates, contributing a sigmoid response to $V_q$ with magnitude depending on the difference in depth.

\begin{equation}
    V_{q,2} = \frac{1}{K} \sum_{v=0}^{K-1} C_{v,p} \frac{1}{1 + \exp \left( {-\lambda_p (D_{v,p} - S^v_q)} \right) }
\end{equation}
Here, $V_{q,2}$ is the free-space violation response for the depth hypothesis at voxel $S_q$ encoded into the third channel of $V_q$. The response values are weighted by the original confidence values $C_{v,p}$ for view $v$ and pixel $p$. The multiplier $\lambda_p$ is the same parameter used in the encoding of occlusions.

\subsection{Evidence Aggregation and Depth Estimation}
\label{sec:method_depth_est}
In order to aggregate neighboring information, we regularize the VCV using a 3D UNet similar to MVSNet \cite{Yao_2018_ECCV}. This includes several layers of 3D convolutions with down-sampling and skip-connections to incorporate global context in the latent space, producing a probability volume $P$. We then apply a soft-argmax operator along the depth dimension. The final fused depth map is generated using the depth-wise expectation of probabilities for each depth hypothesis,

\begin{equation}
    D^f_p = \sum_d S_{p,d} P_{p,d}
\end{equation}
Here, we write $S_{p,d}$ and $P_{p,d}$ using explicit index notation instead of $S_q$ and $P_q$ to clearly indicate the reduction over the depth dimension.

\subsection{Dynamic Depth Search Windows}
\label{sec:method_swe}
As input, our VCV construction process takes a set of depth hypotheses per ray. Instead of a single constant set of hypotheses for all rays, we aim to formulate a hypothesis set per ray that is learned from the data. For the sake of run-time and memory efficiency, it is important to limit the number of depth hypotheses. Therefore, we look to reduce the search space while maintaining a high probability that it encompasses the true depth.

Our Search Window Estimation (SWE) sub-network takes as input the $N$ rendered depth and confidence maps. We compute the mean and standard deviation of both the depth and confidence maps per-pixel. Similar to the formulations of the constraints, we average these metrics over the set of valid inputs $K \leq N$. In order to center the search windows around an initial value $D^{center}$, we use the input confidence values to select the most confident depth estimate from the $N$ input views. See the supplement for the motivation behind this choice.

The input to our search window estimation sub-network is the concatenation of the pixel-wise depth and confidence statistics with $D^{center}$. We run this 5-channel feature map through several 2D convolutional layers, followed by a sigmoid activation function. The output is used for estimating the search window radius,

\begin{equation}
    R_p = r_{min} + r_{max}O_p
\end{equation}
where $R_p$ is the window radius at pixel $p$, $r_{min} = \psi_{min} (b_{max} - b_{min})$ and $r_{max} = \psi_{max} (b_{max} - b_{min})$ are the minimum and maximum allowable bound for the window radius respectively, and $O_p$ is the output of the 2D convolutional network at pixel $p$. The scalars $\psi_{min}$ and $\psi_{max}$ are used to select a percentage of the full input hypothesis range $(b_{max} - b_{min})$ as the minimum and maximum allowable search window radii. These parameters are in place to prevent the network from estimating extreme radius values.

Using this estimated window radius, we can define the depth hypothesis bounds centered around the initial window center estimates.

\begin{align}
    B^{min}_p  &= D^{center}_p - R_p \\
    B^{max}_p   &= D^{center}_p + R_p
\end{align}
Here, $B^{min}_p$ and $B^{max}_p$ are the minimum and maximum depth bounds defining the search window at pixel $p$. We then interpolate between these new bounds to obtain $M$ depth hypotheses, $S_p = [B^{min}_p,...,B^{max}_p] \in \mathbb{R}^{M}$. The new hypothesis volume $S$, with per-pixel hypotheses sets, is then used to build the VCV as described in Section \ref{sec:method_vcv}.

%% file: loss.tex
We train our network in a supervised manner on the output depth and confidence maps of MVS frameworks. We formulate two loss functions, one for each sub-network.

\subsection{Depth Regression Loss}
\label{sec:loss_reg}
We specify the depth regression loss as the $l_1$ loss between the estimated fused depth maps $D^f$ and the ground truth depth maps $D^{gt}$.

\begin{equation}
    L_d = \sum_{p \in \Omega_p} |D^f_p - D^{gt}_p|
\end{equation}
Here, $\Omega_p$ is the set of all valid pixels where ground truth depths 
are available.
This loss is mainly used to supervise the construction of the VCV and the regularization network; however, there are no barriers in place to prevent back-propagation through the SWE sub-network. That being said, it is not sufficient to rely on the regression loss to supervise our SWE sub-network.

\subsection{Depth Search Window Loss}
\label{sec:loss_swe}
In order to supervise the SWE network module, we formulate two objective functions. The first term, named the \textit{coverage loss}, penalizes estimated search windows that do not encompass the ground truth depth.

\begin{equation}
    L_{c} = \sum_{p \in \Omega_p} \frac{|D^{center}_p - D^{gt}_p|}{R_p}
\end{equation}
Using the coverage loss in isolation would not prevent the network from learning to simply maximize the window radius. Therefore, as a regularizing term, we add the magnitude of the window radius to the joint loss function.

\begin{equation}
    L_{r} = \sum_{p \in \Omega_p} R_p
\end{equation}
This term directly penalizes large window radii, guiding the SWE sub-network to produce tight search windows that include the ground truth depth. The formulation of the loss in this manner bears some similarity to the work of Kendall and Gal \cite{kendall2017uncertainties}, with the window radii used as a proxy for uncertainty.

Our total loss is the weighted sum of these three objective functions.

\begin{equation}
    L = \lambda_d L_d + \lambda_c L_c + \lambda_r L_r
\end{equation}

\newcommand\werr{0.30\columnwidth}
\begin{figure}[t]
\footnotesize
    \centering
    \begin{tabular}{ccccc}
        \includegraphics[width=\werr]{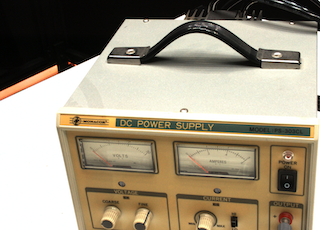} & \hspace{-10pt}
        \includegraphics[width=\werr]{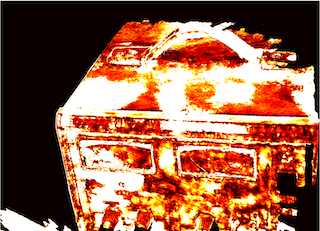} & \hspace{-10pt}
        \includegraphics[width=\werr]{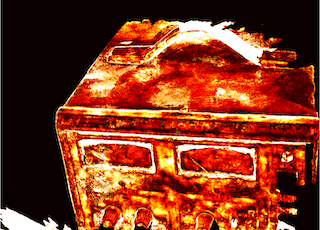} \\
        \includegraphics[width=\werr]{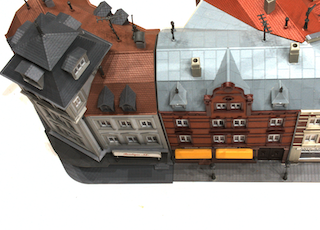} & \hspace{-10pt}
        \includegraphics[width=\werr]{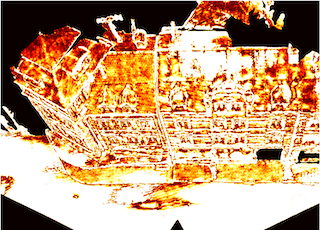} & \hspace{-10pt}
        \includegraphics[width=\werr]{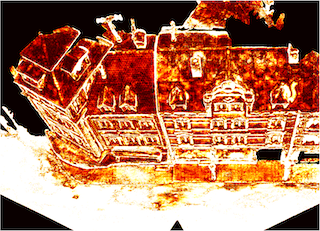} \\
        (c) Image & \hspace{-10pt}
        (b) Input Error & \hspace{-10pt}
        (b) Fused Error \\
    \end{tabular}
    \caption{Error comparison between the input depth maps from NP-CVP-MVSNet \cite{Yang_2022_CVPR} and the fused maps from \network. The errors are encoded as heat maps, with brighter colors corresponding to higher errors. \network{} helps recover from inconsistencies due to texture-less regions, such as the top of the power supply and sections of the roof.}
    \label{fig:error_maps}
\end{figure}

%% file: experiments.tex
\newcommand\wdc{0.30\columnwidth}
\begin{figure}[t]
\footnotesize
    \centering
    \begin{tabular}{ccc}
        \includegraphics[width=\wdc]{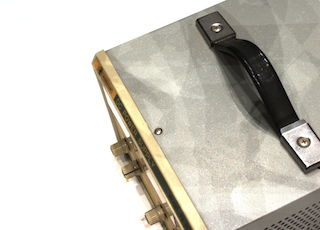} & \hspace{-10pt}
        \includegraphics[width=\wdc]{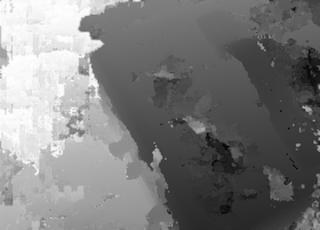} & \hspace{-10pt}
        \includegraphics[width=\wdc]{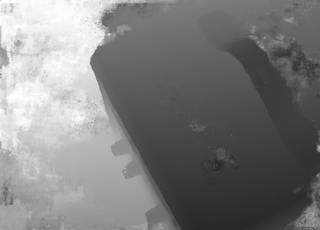} \\
        \includegraphics[width=\wdc]{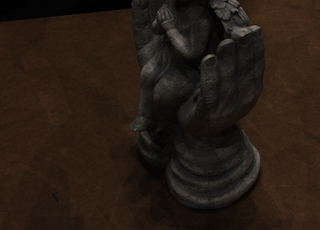} & \hspace{-10pt}
        \includegraphics[width=\wdc]{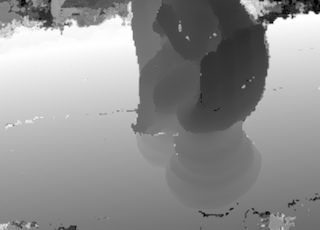} & \hspace{-10pt}
        \includegraphics[width=\wdc]{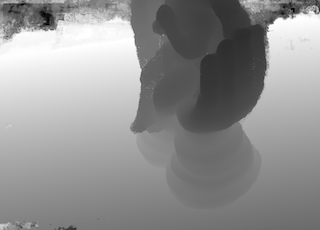} \\
        (a) Image & \hspace{-10pt}
        (b) Input Depth & \hspace{-10pt}
        (c) Fused Depth 
    \end{tabular}
    \caption{Qualitative examples comparing the input depth maps with the fused output depth maps from the DTU \cite{aanaes16} dataset using GBi-Net \cite{Mi_2022_CVPR} as input. The depth maps generated by \network{} improve fine details, texture-less surfaces, and estimates near depth discontinuities.}
    \label{fig:depth_comp}
\end{figure}

\begin{table}[t]
\resizebox{\columnwidth}{!}{%
\begin{tabular}{l|c|cccc}
\hline
    \multirow{2}{*}{Method} &
    \multicolumn{5}{c}{DTU} \\ \cline{2-6}
    &
      \multicolumn{1}{c|}{MAE$\downarrow$} &
      $<0.125\uparrow$ &
      $<0.25\uparrow$ &
      $<0.50\uparrow$ &
      $<1.00\uparrow$ \\ \hline\hline
    MVSNet \cite{Yao_2018_ECCV} &
      9.200 &
      9.55 &
      18.67 &
      34.55 &
      55.35 \\
    + Conventional \cite{Merrell_2007_ICCV} &
      9.050 &
      13.19 &
      25.57 &
      45.34 &
      65.16 \\
    + \network{} &
      \textbf{6.838} &
      \textbf{15.00} &
      \textbf{28.57} &
      \textbf{48.45} &
      \textbf{66.51} \\ \hline
    UCSNet \cite{Cheng_2020_CVPR} &
      12.071 &
      9.99 &
      19.52 &
      35.90 &
      56.24 \\
    + Conventional \cite{Merrell_2007_ICCV} &
      13.633 &
      12.45 &
      24.06 &
      42.59 &
      61.15 \\
    + \network{} &
      \textbf{9.667} &
      \textbf{12.85} &
      \textbf{24.85} &
      \textbf{43.96} &
      \textbf{62.69} \\ \hline
    NP-CVP-MVSNet \cite{Yang_2022_CVPR} &
      12.897 &
      11.76 &
      23.16 &
      42.92 &
      64.27 \\
    + Conventional \cite{Merrell_2007_ICCV} &
      11.933 &
      12.79 &
      25.35 &
      47.42 &
      68.94 \\
    + \network{} &
      \textbf{8.566} &
      \textbf{16.47} &
      \textbf{31.49} &
      \textbf{53.06} &
      \textbf{70.58} \\ \hline
    GBi-Net \cite{Mi_2022_CVPR} &
      5.845 &
      12.77 &
      24.89 &
      45.10 &
      65.94 \\
    + Conventional \cite{Merrell_2007_ICCV} &
      5.009 &
      17.30 &
      32.93 &
      55.52 &
      73.66 \\
    + \network{} &
      \textbf{4.196} &
      \textbf{18.41} &
      \textbf{34.79} &
      \textbf{57.50} &
      \textbf{74.66} \\ \hline
\end{tabular}%
}
\caption{Quantitative comparison of the 2D depth map errors on the evaluation set of DTU \cite{aanaes16} benchmark. All threshold values are measured in \textit{mm}. Conventional fusion improves almost all inputs, even those from recent state-of-the-art methods, in terms of average error over all pixels with ground truth and also by increasing the number of inliers for each threshold. Learned fusion via \network{} leads to even larger improvements in \textit{all} cases.
}
\label{tab:dtu-2d-error}
\vspace{-5pt}
\end{table}

\subsection{Datasets}
\noindent
\textbf{DTU}
The DTU dataset \cite{aanaes16} is an indoor dataset that contains images of 124 scenes taken from a camera mounted on an industrial robot arm. All scenes share the same camera trajectories, with ground-truth point clouds captured via a structured light scanner. We follow the training, validation, and evaluation split used by Yao et al. \cite{Yao_2018_ECCV}.

\noindent
\textbf{Tanks \& Temples}
The Tanks \& Temples dataset \cite{Knapitsch_2017_TNT} is a large-scale, mostly outdoor dataset containing video sequences of challenging scenes. The dataset is divided into a training set and two evaluations sets; intermediate and advanced.

\noindent
\textbf{BlendedMVS}
The BlendedMVS \cite{yao2020blendedmvs} dataset is a large-scale, synthetic dataset containing images processed by blending original images with rendered images from each scene mesh. The dataset is split into training and validation sets, containing 106 and 7 scenes respectively.

\subsection{Implementation Details}
\label{sec:implementation}
\noindent
\textbf{Training Details}
 We implement the model with PyTorch \cite{paszke2019pytorch} and train on the output depth and confidence maps of the DTU \cite{aanaes16} dataset from several deep MVS methods, separately. For improved generalization, we follow the robust training strategy used by PatchmatchNet \cite{Wang_2021_CVPR}, in which we randomly choose $N-1$ of the $10$ best source views to use as training for a given reference view. We train on an NVIDIA RTX A6000 GPU for 30 epochs. The model has approximately 300,000 parameters and training takes 1 hour per epoch for the high resolution data. We use the Adam optimizer with a learning rate of $0.0003$ using an exponential decay of $0.95$ every $2$ epochs. For additional model parameters, please see the supplement.

\subsection{Evaluation}
\noindent
\textbf{Metrics}
As our method is focused on generating depth and confidence maps, we mainly focus our evaluation on 2D metrics. For depth map evaluation, we report the \textit{mean absolute error (MAE)} between the estimated and ground truth depth maps. We also report the percentage of pixels with depth estimates within several error thresholds. We also present 3D metrics on output point clouds generated from the fused depth maps. We evaluate our point clouds on the DTU benchmark \cite{aanaes16}, measuring \textit{accuracy}, \textit{completeness}, and \textit{overall} scores. Accuracy is the mean distance between every point in the estimated point cloud to the closest point in the ground truth model and completeness is the mean distance between every point in the ground truth point cloud to the closest point in the estimated model. The overall score is the average of these metrics. We show a variation of these metrics when comparing to DeFuSR \cite{Donne_2019_CVPR} following the evaluations performed in their work. Donn{\'e} and Geiger \cite{Donne_2019_CVPR} report the Chamfer distances as the percentage of points within a threshold of $\tau = 2.0mm$. We also evaluate our point clouds on the Tanks \& Temples benchmark. We report the \textit{f-score} for each scene, as well as the mean f-score for all scenes.

\begin{table}[t]
\centering
\resizebox{0.99\columnwidth}{!}{%
\begin{tabular}{l|ccc|ccc}
\hline
    \multirow{2}{*}{Method} &
    \multicolumn{3}{c}{DTU [Sparse] (mm) $\downarrow$} &
    \multicolumn{3}{c}{DTU [Dense] (mm) $\downarrow$} \\ \cline{2-7}
    &
    Acc. &
    Comp. &
    \textbf{Overall} &
    Acc. &
    Comp. &
    \textbf{Overall} \\ \hline\hline
    \textbf{{\ul MVSNet \cite{Yao_2018_ECCV}}} &&&&&& \\
    + Gipuma \cite{Galliani_2015_ICCV} &
      \textbf{0.396} &
      0.527 &
      0.462 &
      0.419 &
      0.383 &
      0.401 \\
    + \network{} &
      0.432 &
      \textbf{0.390} &
      \textbf{0.411} &
      \textbf{0.388} &
      \textbf{0.349} &
      \textbf{0.368} \\ \hline
    \textbf{{\ul UCSNet \cite{Cheng_2020_CVPR}}} &&&&&& \\
    + Gipuma \cite{Galliani_2015_ICCV} &
      \textbf{0.338} &
      0.349 &
      0.344 &
      0.320 &
      \textbf{0.261} &
      0.290 \\
    + \network{} &
      0.354 &
      \textbf{0.329} &
      \textbf{0.342} &
      \textbf{0.265} &
      0.276 &
      \textbf{0.270} \\ \hline
    \textbf{{\ul NP-CVP-MVSNet \cite{Yang_2022_CVPR}}}  &&&&&& \\
    + Gipuma \cite{Galliani_2015_ICCV} &
      0.356 &
      \textbf{0.275} &
      0.316 &
      0.288 &
      0.194 &
      0.241 \\
    + \network{} &
      \textbf{0.337} &
      0.277 &
      \textbf{0.307} &
      \textbf{0.256} &
      \textbf{0.181} &
      \textbf{0.219} \\ \hline
    \textbf{{\ul GBi-Net \cite{Mi_2022_CVPR}}}  &&&&&& \\
    + $\sim$ COLMAP \cite{schonberger2016eccv} &
      0.315 &
      \textbf{0.262} &
      \textbf{0.289} &
      0.254 &
      \textbf{0.173} &
      0.214 \\
    + \network{} &
      \textbf{0.310} &
      0.274 &
      0.292 &
      \textbf{0.227} &
      0.180 &
      \textbf{0.204} \\ \hline
\end{tabular}%
}
\caption{Chamfer distances (lower is better) of the final fused point clouds from the evaluation set of DTU \cite{aanaes16} benchmark. We evaluate the final models using the official script that enforces a sparse minimum point-spacing of 0.2mm (left). Since the errors are approaching this threshold, we also evaluate the models enforcing a dense minimum point-spacing of 0.03mm (right). MVSNet \cite{Yao_2018_ECCV}, UCSNet \cite{Cheng_2020_CVPR}, and NP-CVP-MVSNet \cite{Yang_2022_CVPR} use Gipuma \cite{Galliani_2015_ICCV} to fuse depth estimates into a final 3D model. GBi-Net \cite{Mi_2022_CVPR} uses an adaptation of the fusion approach of COLMAP, in which geometric and photometric filters are used to filter and average consistent depth estimates across views.}
\label{tab:dtu-results}
\vspace{-10pt}
\end{table}

\noindent
\textbf{MVS Baselines} We compare the results of applying \network{} on the outputs of MVSNet \cite{Yao_2018_ECCV}, UCSNet \cite{Cheng_2020_CVPR} as a representative multi-resolution algorithm and two state-of-the-art methods, NP-CVP-MVSNet \cite{Yang_2022_CVPR} and GBi-Net \cite{Mi_2022_CVPR}.

\noindent
\textbf{Fusion Baselines} We compare the results of \network{} with the conventional fusion approach of Merrell et al. \cite{Merrell_2007_ICCV} for 2D evaluations, and Gipuma \cite{Galliani_2015_ICCV} for 3D evaluations, since Gipuma is the method of choice by \textit{state-of-the-art} MVS frameworks to produce final 3D models. We also provide comparisons to the learning-based fusion method, DeFuSR \cite{Donne_2019_CVPR}. Methods operating on implicit TSDF volumes, such as VolumeFusion \cite{Choe_2021_ICCV}, RoutedFusion \cite{Weder_2020_CVPR}, and NeuralFusion \cite{Weder_2021_CVPR} are not included in our evaluations, since they are better suited for reconstructing closed, watertight objects. These papers do not provide any quantitative evaluations on DTU or Tanks \& Temples, with NeuralFusion presenting qualitative-only results on select scenes from Tanks \& Temples.

\begin{table}[t]
\centering
\resizebox{0.85\columnwidth}{!}{%
\begin{tabular}{l|ccc}
\hline
    \multirow{2}{*}{Method} &
    \multicolumn{3}{c}{DTU (full) $\uparrow$} \\ \cline{2-4}
    &
    Acc. (\%) &
    Comp. (\%) &
    Mean (\%) \\ \hline\hline
    MVSNet \cite{Yao_2018_ECCV} &
      88 &
      66 &
      77 \\
    + DeFuSR \cite{Donne_2019_CVPR} &
      86 &
      65 &
      76 \\   
    + \network{} &
      \textbf{98} &
      \textbf{98} &
      \textbf{98} \\ \hline
\end{tabular}%
}
\caption{Chamfer distances (lower is better) of the final 3D models of MVSNet \cite{Yao_2018_ECCV} using DeFuSR \cite{Donne_2019_CVPR} and \network{} for fusion. Here, accuracy and completeness are reported as the percentage of points with accuracy and completeness scores within $\tau = 2.0mm$.}
\label{tab:defusr_comp}
\vspace{-5pt}
\end{table}

\newcommand\wtp{0.15\textwidth}
\begin{figure*}[t]
\footnotesize
    \centering
    \begin{tabular}{cccccc}
        \includegraphics[width=\wtp]{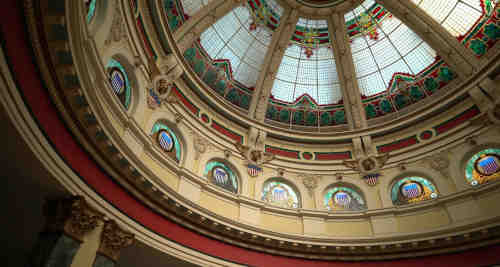} & \hspace{-10pt}
        \includegraphics[width=\wtp]{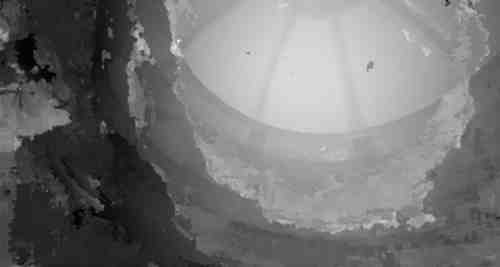} & \hspace{-10pt}
        \includegraphics[width=\wtp]{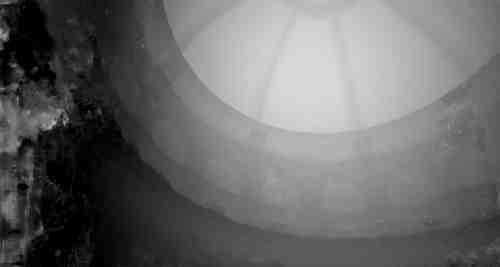} & \hspace{-5pt}
        \includegraphics[width=\wtp]{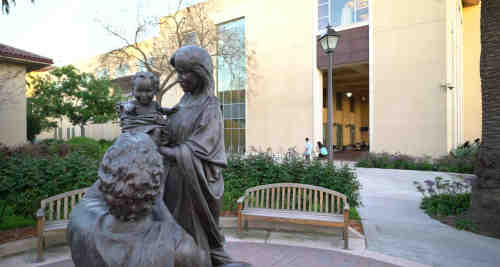} & \hspace{-10pt}
        \includegraphics[width=\wtp]{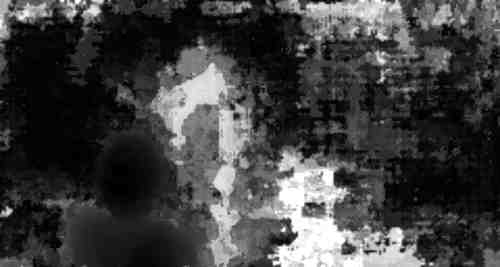} & \hspace{-10pt}
        \includegraphics[width=\wtp]{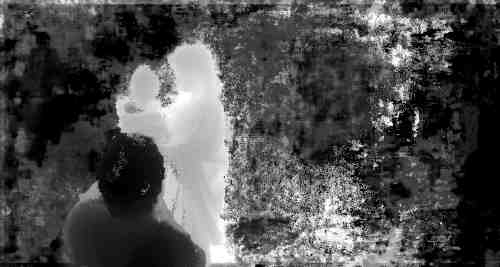} \\
        \includegraphics[width=\wtp]{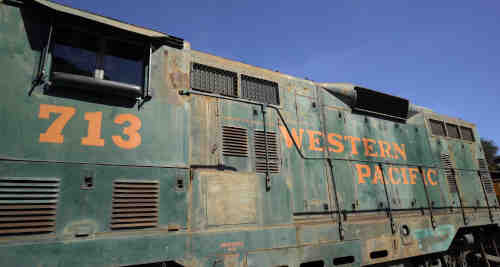} & \hspace{-10pt}
        \includegraphics[width=\wtp]{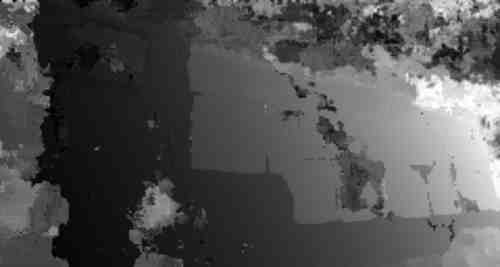} & \hspace{-10pt}
        \includegraphics[width=\wtp]{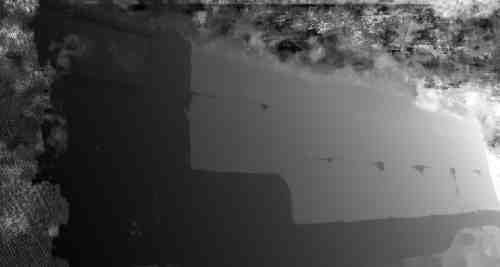} & \hspace{-5pt}
        \includegraphics[width=\wtp]{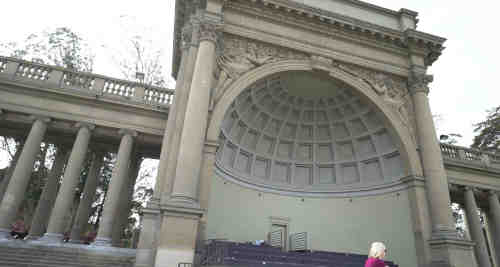} & \hspace{-10pt}
        \includegraphics[width=\wtp]{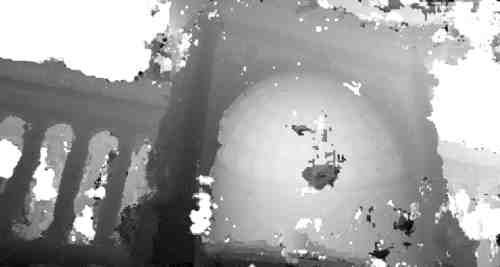} & \hspace{-10pt}
        \includegraphics[width=\wtp]{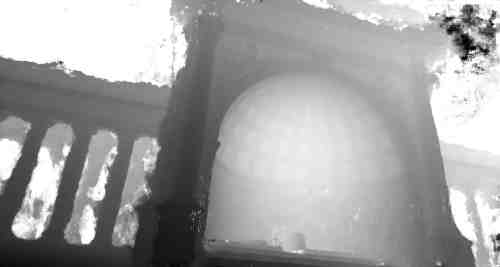} \\
        (a) Image & \hspace{-10pt}
        (b) Input Depth & \hspace{-10pt}
        (c) Fused Depth & \hspace{-5pt}
        (d) Image & \hspace{-10pt}
        (e) Input Depth & \hspace{-10pt}
        (f) Fused Depth
    \end{tabular}
    \caption{Qualitative comparison of depth maps for scenes from the Tanks \& Temples benchmark \cite{Knapitsch_2017_TNT}  using GBi-Net \cite{Mi_2022_CVPR} as input.}
    \label{fig:tnt_qual}
\end{figure*}

\begin{table*}[t]
\centering
\resizebox{0.8\textwidth}{!}{%
\begin{tabular}{l|ccccccccc}
\hline
\multirow{2}{*}{Method} &
  \multicolumn{9}{c}{intermediate $\uparrow$} \\ \cline{2-10} 
 &
  \textbf{Mean} &
  Fam. &
  Franc. &
  Horse &
  Light. &
  M60 &
  Pan. &
  Play. &
  Train \\ \hline \hline

\textbf{{\ul UCSNet \cite{Cheng_2020_CVPR}}} &&&&&&&&& \\
+ Gipuma \cite{Galliani_2015_ICCV} &
  54.83 &
  \textbf{76.09} &
  53.16 &
  43.03 &
  54.00 &
  55.60 &
  \textbf{51.49} &
  \textbf{57.38} &
  \textbf{47.89} \\
+ \network{} &
  \textbf{55.03} &
  75.64 &
  \textbf{57.60} &
  \textbf{46.03} &
  \textbf{54.35} &
  \textbf{55.78} &
  49.42 &
  56.02 &
  45.37 \\ \hline
\textbf{{\ul GBi-Net \cite{Mi_2022_CVPR}}} &&&&&&&&& \\
+ $\sim$ COLMAP \cite{schonberger2016eccv} &
 \textbf{ 61.42} &
  \textbf{79.77} &
 \textbf{ 67.69} &
  \textbf{51.81} &
  \textbf{61.25} &
  \textbf{60.37} &
  \textbf{55.87} &
  \textbf{60.67} &
  \textbf{53.89} \\
+ \network{} &
  59.08 &
  78.92 &
  65.23 &
  49.96 &
  59.16 &
  57.08 &
  53.13 &
  58.58 &
  50.61 \\ \hline
\end{tabular}%
}
\caption{F-score (higher is better) of the final fused point clouds from the evaluation sets of the Tanks \& Temples \cite{Knapitsch_2017_TNT} benchmark. The best results between the baseline and \network{} are marked as bold.}
\label{tab:tnt-results}
\end{table*}

\begin{table}[t]
\resizebox{\columnwidth}{!}{%
\begin{tabular}{l|c|ccc}
\hline
    \multirow{2}{*}{Method} &
    \multicolumn{4}{c}{Tanks \& Temples} \\ \cline{2-5}
    &
      \multicolumn{1}{c|}{MAE$\downarrow$} &
      $<\tau\uparrow$ &
      $<2\tau\uparrow$ &
      $<4\tau\uparrow$ \\ \hline\hline
    UCSNet \cite{Cheng_2020_CVPR} &
      0.175 &
      11.83 &
      19.69 &
      \textbf{30.17} \\
    UCSNet + \network{} &
      \textbf{0.167} &
      \textbf{12.18} &
      \textbf{19.75} &
      29.84 \\ \hline
    NP-CVP-MVSNet \cite{Yang_2022_CVPR} &
      0.177 &
      15.49 &
      \textbf{25.16} &
      37.38\\
    NP-CVP-MVSNet + \network{} &
      \textbf{0.155} &
      \textbf{15.68} &
      25.13 &
      \textbf{37.57} \\ \hline
    GBi-Net \cite{Mi_2022_CVPR} &
      \textbf{0.240} &
      12.02 &
      19.51 &
      29.24 \\
    GBi-Net + \network{} &
      0.243 &
      \textbf{12.88} &
      \textbf{20.57} &
      \textbf{30.22} \\ \hline
\end{tabular}%
}
\caption{Quantitative comparison of depth map errors on the training set of Tanks \& Temples \cite{Knapitsch_2017_TNT}. All methods have been trained on DTU. The threshold value $\tau$ is selected per-scene and is derived from the thresholds provided by the benchmark.}
\label{tab:tnt-2d-error}
\vspace{-10pt}
\end{table}

\noindent
\textbf{Evaluation on DTU Dataset}
We first compute ground truth depth maps for DTU in the same manner as MVSNet \cite{Yao_2018_ECCV}. Specifically, we run screened Poisson surface reconstruction (SPSR) \cite{Kazhdan_2013_TOG} on the provided ground truth point clouds for each scene and produce a watertight mesh. We then render this mesh into all cameras to obtain ground truth depth maps. To produce our final point clouds, we use heuristic filtering, similar to the post-processing presented in GBi-Net. We first filter out depth estimates that have a confidence value below a threshold. We then project each estimate into neighboring views, using the depth estimates in each view to reproject back to the reference view, measuring the pixel reprojection error and filtering out estimates whose error is above a threshold. 

Table \ref{tab:dtu-2d-error} shows a comparison of the depth map errors between all baseline methods and \network{}. Observing the fused depth map errors, we can see that even using the low resolution inputs of MVSNet, \network{} can generate depth maps with a lower MAE than UCSNet and NP-CVP-MVSNet. Additionally, \network{} produces depth maps with more inliers at all threshold values compared to the input depth maps generated by all baseline methods. A comparison of error maps is shown in Figure \ref{fig:error_maps}. Qualitative depth map results can be seen in Figure \ref{fig:depth_comp}. We can observe that \network{} removes much of the noise in the input depth maps, while producing better estimates near depth discontinuities. In Table \ref{tab:dtu-results}, we evaluate the final 3D models of all MVS baselines and compare the fusion choice from each method to \network{}. \network{} shows clear improvements in the \textit{overall} results in both Sparse and Dense evaluation scenarios. In the case of GBi-Net, the improvements realized by \network{} are more noticeable without the sampling procedure used in the Sparse evaluation. DeFuSR \cite{Donne_2019_CVPR} provides results evaluating fusion of COLMAP \cite{schonberger2016eccv} and MVSNet \cite{Yao_2018_ECCV} inputs on DTU.
We provide a comparison according to the evaluation protocol used in \cite{Donne_2019_CVPR} in Table \ref{tab:defusr_comp}. The threshold used by DeFuSR is $\tau = 2.0mm$, which is quite large. \network{} outperforms DeFuSR by a substantial margin, which is expected as the authors state they are not able to refine the MVSNet inputs much.

\noindent
\textbf{Evaluation on Tanks \& Temples Dataset}
We use the model trained on the DTU output depth and confidence maps of each network without any fine-tuning for evaluation. In order to evaluate the depth maps on Tanks \& Temples, we use the training set and the provided ground truth point clouds, computing ground truth depth maps the same way they are computed for DTU. Table \ref{tab:tnt-2d-error} shows the depth map errors between each baseline and \network{}. The fused depth maps are more accurate overall for UCSNet and NP-CVP-MVSNet. For GBi-Net, we show improved accuracy for estimates within the error thresholds. See Figure \ref{fig:tnt_qual} for a qualitative comparison of depth maps. Table \ref{tab:tnt-results} shows the f-scores for the final point clouds on the Tanks \& Temples intermediate set. We show comparable results to both input MVS baselines. We provide the \textit{precision} and \textit{recall} split for each method in the supplement.

\noindent
\textbf{Additional Experiments}
We provide results on the validation set of the BlendedMVS \cite{yao2020blendedmvs} dataset in the supplement. Using the outputs of GBi-Net trained on the BlendedMVS training set and the \network{} model trained on DTU without any fine-tuning, \network{} produces higher quality depth maps for all scenes, with a mean MAE of \textit{0.288} compared to \textit{0.319} for GBi-Net. We also provide evaluations of the output confidence maps, reporting the \textit{AUC} of all methods. Using GBi-Net as input, the AUC of \network{} is \textit{2.480} compared to \textit{3.690} for GBi-Net. We show several ablation studies in the supplement, testing the individual contributions of different aspects of the network architecture. Specifically, we evaluate the contributions of the visibility constraints, as well as the efficiency gains of the SWE sub-network. As detailed in the supplement, introducing the SWE sub-network results in $8.5 \times$ memory and $9 \times$ run-time efficiency gains, as well as a $20\%$ decrease in MAE.

%% file: conclusion.tex
We have presented an end-to-end depth map fusion network that leverages long-range visibility constraints encoded into a learnable pipeline. Our method improves input depth and confidence maps generated by MVS networks, integrating multi-view consensus and inconsistency measures. We also present a novel depth search space refinement sub-network that estimates a narrow search window along each ray to increase memory and run-time efficiency, as well as allow for high resolution depth estimation near surfaces. The combination of these concepts is able to obtain fused depth maps that are quantitatively and qualitatively much better than the inputs. While the depth map fusion in our work is end-to-end, merging the depth estimates into a unified point cloud remains a heuristic-driven process. We aim to incorporate a more principled point cloud reconstruction procedure from a collection of depth maps in future work. We also aim to explore the generalization ability of learning-based fusion.

%% file: supplement.tex
\setcounter{figure}{0}
\setcounter{table}{0}
\setcounter{section}{0}
\renewcommand\thefigure{S.\arabic{figure}}  
\renewcommand\thetable{S.\arabic{table}}
\renewcommand\thesection{S.\arabic{section}}

\section*{Supplementary Material}
In this document, we provide additional implementation details (Section \ref{sec:sup_impl}), formulations for learnable geometric constraints (Section \ref{sec:sup_hyp}), details on confidence computation and evaluation (Section \ref{sec:sup_conf}), ablation studies evaluating architecture contributions (Section \ref{sec:sup_ablations}), further quantitative evaluations (Section \ref{sec:sup_quant}), and additional qualitative comparisons and examples (Section \ref{sec:sup_qual}).

\section{Implementation Details}
\label{sec:sup_impl}
During training and inference, we set the number of input views for fusion to $N=5$ and the number of depth planes to $M=8$ for input depth maps from all methods. For the DTU \cite{aanaes16} dataset, we use the maximum output resolution of each method as the input for \network{}. Specifically, we use $400 \times 296$ for MVSNet, $1600 \times 1184$ for UCSNet, and $1600 \times 1152$ for NP-CVP-MVSNet and GBi-Net. For training, we scale the input by a factor of $0.5$, with the exception of MVSNet, for which we train at the full resolution. For Tanks \& Temples \cite{Knapitsch_2017_TNT}, we use input resolutions of $1920 \times 1056$ for UCSNet and $1920 \times 1024$ for both NP-CVP-MVSNet and GBi-Net. As the minimum and maximum allowable search window radii, we use $\psi_{min} = 0.005$ and $\psi_{max} = 0.50$. The terms of the loss are weighed by $\lambda_d=0.5$, $\lambda_c=20.0$, and $\lambda_r=0.5$.

\section{Hyper-parameters}
\label{sec:sup_hyp}
We define two parameters used in the formulations of the geometric constraints. For support, we use $\sigma_p$ to determine the sharpness of the support response boundary.

\begin{equation}
    \sigma_p = \gamma_{\sigma} \frac{(B^{max}_p - B^{min}_p)}{M (b_{max} - b_{min})}
\end{equation}
Here, $\gamma_{\sigma}$ is a learned hyper-parameter, $B^{min}_p$ and $B^{max}_p$ are the minimum and maximum depth bounds per pixel, $M$ is the number of depth hypotheses, and $b_{min}$ and $b_{max}$ are the overall minimum and maximum depth bounds that are given as input for the current reference view. Lower values of $\gamma_{\sigma}$ correspond to a tighter support window. Since this is a function of the per-pixel depth bounds, support adapts to the confidence at each pixel.

\begin{figure}[t]
\footnotesize
    \centering
    \begin{tabular}{cc}
        \includegraphics[width=0.48\columnwidth]{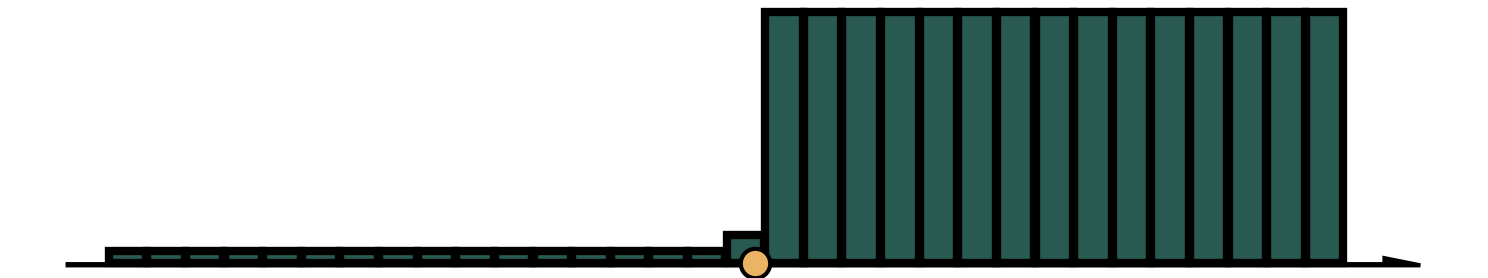} &
        \includegraphics[width=0.48\columnwidth]{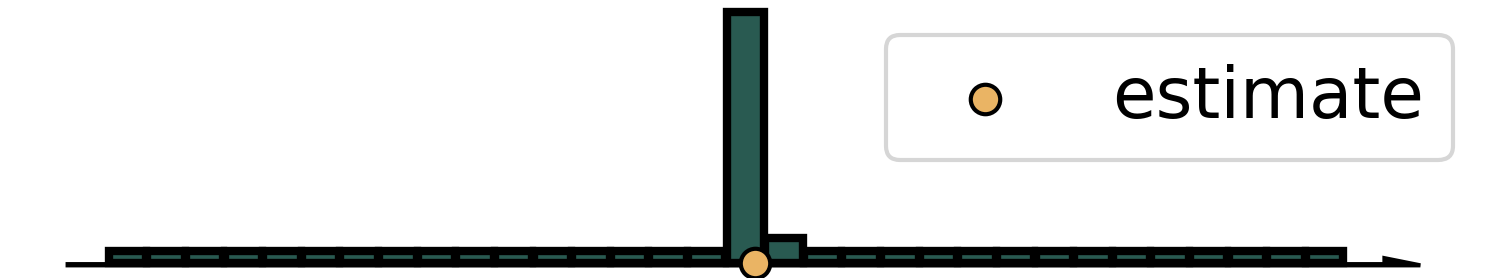} \\
        \includegraphics[width=0.48\columnwidth]{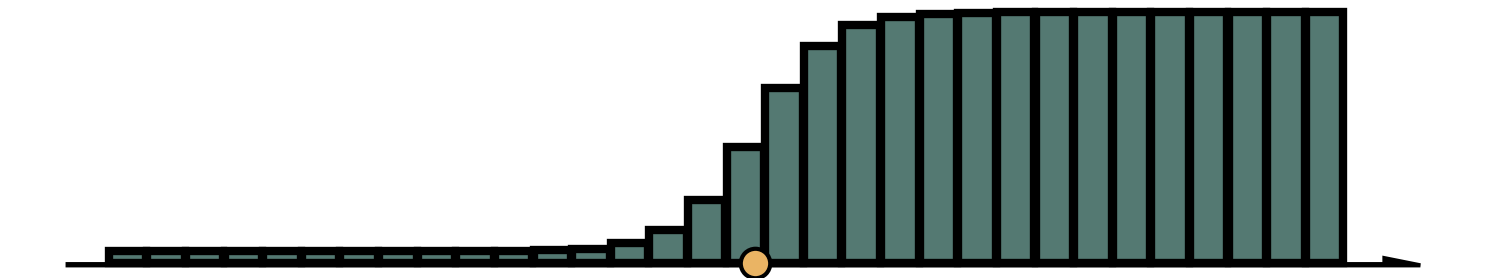} &
        \includegraphics[width=0.48\columnwidth]{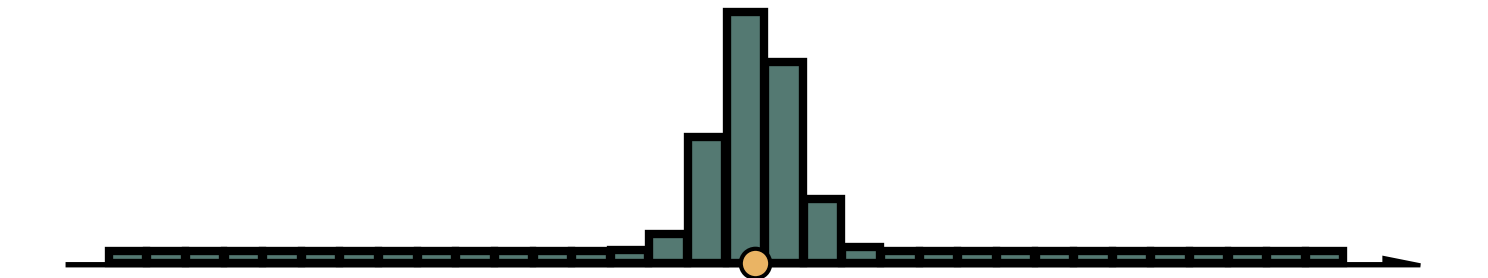} \\
        \includegraphics[width=0.48\columnwidth]{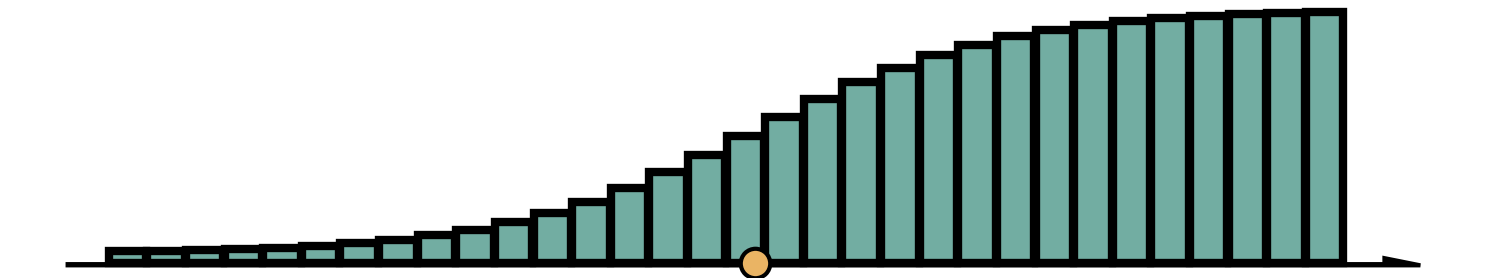} &
        \includegraphics[width=0.48\columnwidth]{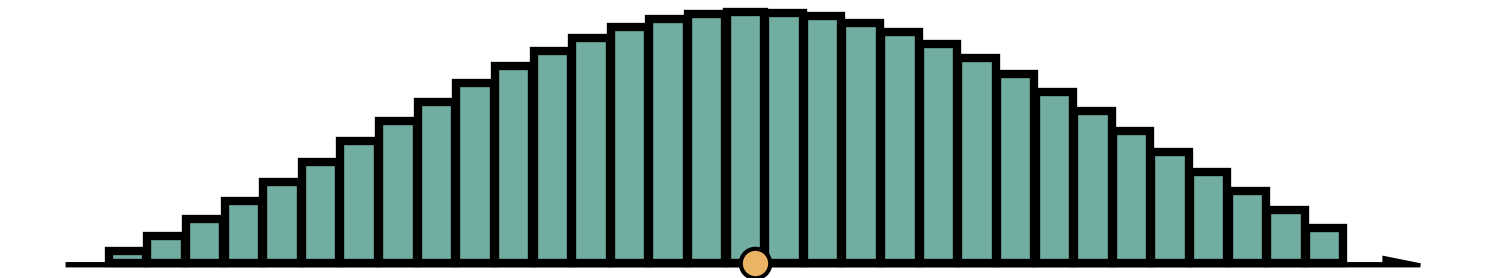} \\
        (a) Occlusions & (b) Support
    \end{tabular}
    \caption{Visualization of the effects of modifying (a), the multiplier used in the sigmoid and (b), the sigma used in the Gaussian. The \textit{response boundary} is the region around the current depth estimate where high response values transition to low response values. Decreasing the multiplier for occlusions (as well as free-space violations) (a; top$\rightarrow$bottom) causes the sigmoid response boundary to soften around the current estimated depth. Increasing the sigma for support (b; top$\rightarrow$bottom) causes the support response boundary to soften around the current estimated depth.}
    \label{fig:response_boundary}
\end{figure}

For occlusions and free-space violations, we use $\lambda_p$ to determine the sharpness of the sigmoid response boundary.

\begin{equation}
    \lambda_p = \gamma_{\lambda} \frac{M (b_{max} - b_{min})}{(B^{max}_p - B^{min}_p)}
\end{equation}
Here, $\gamma_{\lambda}$ is a learned hyper-parameter. Lower values of $\gamma_{\lambda}$ correspond to a softer sigmoid response boundary. Occlusions and free-space violations also adapt to the confidence at each pixel. See Figure \ref{fig:response_boundary} for a visualization of the response boundary.

\begin{table}[t]
\centering
\resizebox{\columnwidth}{!}{%
\begin{tabular}{l|c|c}
\hline
    \multirow{2}{*}{Method} &
    \multicolumn{2}{c}{AUC$\downarrow$} \\ \cline{2-3}
    &
    DTU &
    Tanks \& Temples \\ \hline\hline
    MVSNet \cite{Yao_2018_ECCV} &
      6.12 &
      - \\
    MVSNet + \network &
      \textbf{4.08} &
      - \\ \hline
    UCSNet \cite{Cheng_2020_CVPR} &
      20.29&
      0.394  \\
    UCSNet + \network &
      \textbf{4.96}&
      \textbf{0.208}  \\ \hline
    NP-CVP-MVSNet \cite{Yang_2022_CVPR} &
      8.28 &
      0.212 \\
    NP-CVP-MVSNet + \network &
      \textbf{4.35} &
      \textbf{0.154} \\ \hline
    GBi-Net \cite{Mi_2022_CVPR} &
      3.69 &
      0.464 \\
    GBi-Net + \network &
      \textbf{2.48} &
      \textbf{0.448} \\ \hline
\end{tabular}%
}
\caption{Quantitative comparison of the average AUC (lower is better) between \network{} and each baseline method on the DTU \cite{aanaes16} evaluation set and the Tanks \& Temples \cite{Knapitsch_2017_TNT} training set.}
\label{tab:auc}
\end{table}

\newcommand\swcc{0.30\columnwidth}
\begin{figure}[t]
\footnotesize
    \centering
    \begin{tabular}{ccc}
        \includegraphics[width=\swcc]{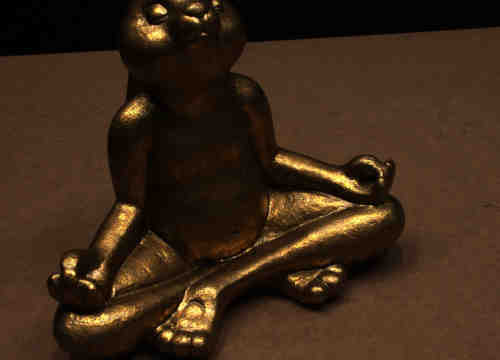} & \hspace{-10pt}
        \includegraphics[width=\swcc]{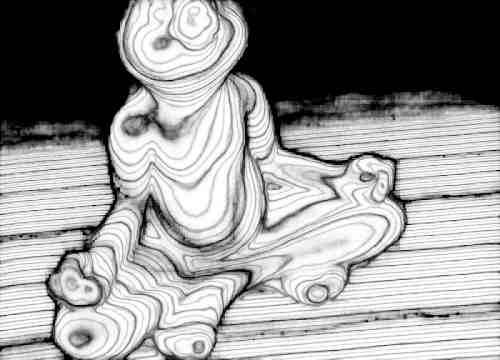} & \hspace{-10pt}
        \includegraphics[width=\swcc]{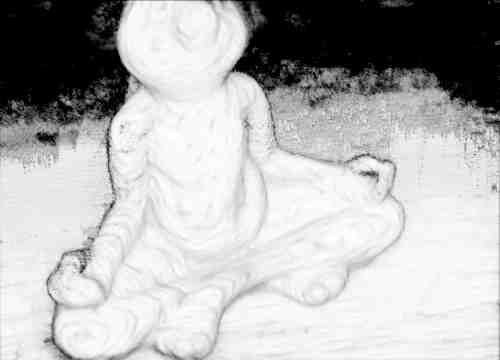} \\
        \includegraphics[width=\swcc]{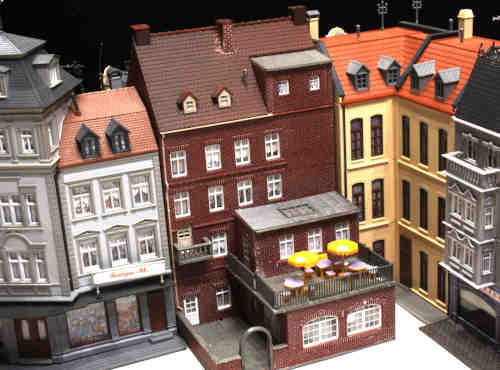} & \hspace{-10pt}
        \includegraphics[width=\swcc]{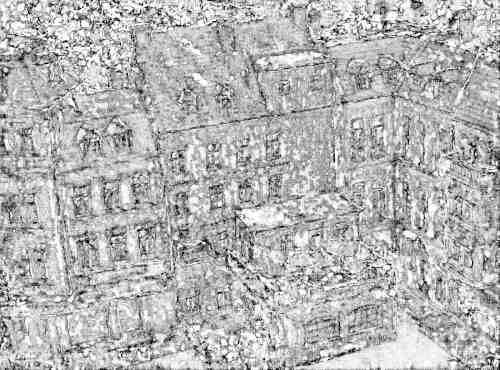} & \hspace{-10pt}
        \includegraphics[width=\swcc]{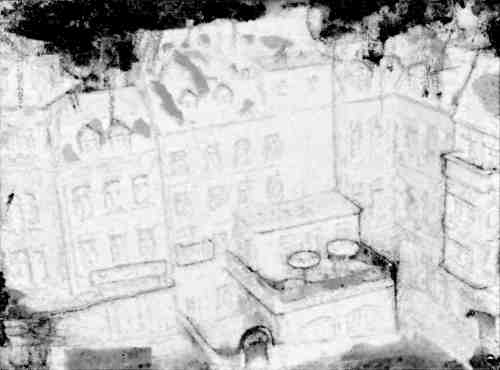} \\
        (c) Image & \hspace{-10pt}
        (d) Input Confidence & \hspace{-10pt}
        (c) Fused Confidence\\
    \end{tabular}
    \caption{Comparison between the input confidence maps with the fused confidence maps on scenes from the DTU \cite{aanaes16} benchmark using GBi-Net \cite{Mi_2022_CVPR} (top), and UCSNet \cite{Cheng_2020_CVPR} (bottom) as input. (Darker pixel value corresponds to lower confidence.)}
    \label{fig:conf_comp}
\end{figure}

\begin{table}[t]
\centering
\begin{tabular}{l|c}
\hline
    \multirow{2}{*}{Method} &
    \multicolumn{1}{c}{DTU} \\ \cline{2-2}
    &
      \multicolumn{1}{c}{MAE$\downarrow$} \\ \hline\hline
    GBi-Net \cite{Mi_2022_CVPR} &
      5.845 \\
    + \network{} [brute-force] &
      5.344 \\
    + \network{} [sup] &
      4.813 \\
    + \network{} [fsv+occ] &
      4.477 \\
    + \network{} &
      \textbf{4.196} \\ \hline
\end{tabular}%
\caption{Ablation study evaluating the contribution from different aspects of the network architecture. [brute-force] indicates the network is trained using the entire search space (using 128 depth planes) without the SWE sub-network. [sup] and [fsv+occ] indicate that the network only utilizes the support channel, or only utilizes the free-space violation and occlusion channels, respectively.}
\label{tab:dtu-2d-ablations}
\end{table}

\section{Confidence Estimates}
\label{sec:sup_conf}
To evaluate the quality of confidence estimates, we report the \textit{area under the curve (AUC)}. The AUC is the area under the ROC curve, which maps the error rate for the depth map as a function of density based on the sorted confidence values \cite{hu2012quantitative}. We first sort the estimated depths according to confidence, and form the sparsification curve of MAE vs. depth map density by progressively dropping the least confident depths \cite{hu2012quantitative} to obtain depth maps of lower density, and presumably lower error. Small area under the sparsification curve indicates that confidence has been estimated well and can be used to rank depth estimates accurately.

The fused confidence maps are directly computed from the estimated search window radius. After we normalize the per-pixel window radii from 0 to 1, the confidence value at each pixel is $C^f_p = 1-R_p$. Intuitively, a larger estimated radius for a given pixel should indicate lower confidence in the final depth estimate. This relationship is also reflected and enforced in our loss function. To compare the quality of the confidence maps, we report the AUC for all methods in Table \ref{tab:auc}. The output confidence values after fusion prove to be more reliable estimates of confidence. Qualitative confidence map results can be seen in Figure \ref{fig:conf_comp}.

\begin{table}[t]
\centering
\resizebox{\columnwidth}{!}{%
\begin{tabular}{l|cc}
\hline
    Method &
    \network{} [brute-force] &
    \network{} [swe] \\ \hline\hline
    Memory(GB) &
    37.734 &
    \textbf{4.439} \\
    Parameters &
    \textbf{289,587} &
    297,429 \\
    Inference Time(s) &
      22.95 &
      \textbf{2.51} \\ \hline
\end{tabular}%
}
\caption{Ablation study between the brute-force approach and the SWE sub-network. We use 192 depth planes for the brute-force approach, with every pixel having the same depth bounds given as input by the dataset. For the SWE sub-network, we use 8 depth planes with each depth bound estimated per-pixel. We can observe significant memory and run-time improvements, with minimal additional model parameters.}
\label{tab:swe-ablations}
\end{table}

\begin{table}[t]
\centering
\begin{tabular}{l|c}
\hline
    \multirow{2}{*}{Method} &
    \multicolumn{1}{c}{DTU} \\ \cline{2-2}
    &
      \multicolumn{1}{c}{AUC$\downarrow$} \\ \hline\hline
    \network{} [pv] &
      7.66 \\
    \network{} [pv+swe] &
      6.06 \\
    \network{} [swe] &
      \textbf{2.48} \\ \hline
\end{tabular}%
\caption{Ablation study on the confidence map computation. We evaluate using only the output probability volume, as well as the incorporated and stand-alone radius outputs from the SWE sub-network (\textit{pv: probability volume; swe: search window estimate}).}
\label{tab:auc-ablations}
\end{table}

\begin{table}[b]
\resizebox{\columnwidth}{!}{%
\begin{tabular}{l|ccc}
\hline
    \multirow{2}{*}{Method} &
    \multicolumn{3}{c}{Tanks \& Temples} \\ \cline{2-4}
    &
      Precision $\uparrow$ &
      Recall $\uparrow$ & 
      F-Score $\uparrow$ \\ \hline\hline
    \textbf{{\ul UCSNet \cite{Cheng_2020_CVPR}}} &&& \\
    + Gipuma \cite{Galliani_2015_ICCV} &
      46.66 &
      \textbf{70.34} &
      54.83 \\
    UCSNet + \network{} &
      \textbf{47.08} &
      68.64 &
      \textbf{55.03} \\ \hline
    \textbf{{\ul GBi-Net \cite{Mi_2022_CVPR}}} &&& \\
    + Gipuma \cite{Galliani_2015_ICCV} &
      \textbf{54.49} &
      71.25 &
      \textbf{61.42} \\
    + \network{} &
      50.16 &
      \textbf{73.08} &
      59.08 \\ \hline
\end{tabular}%
}
\caption{Precision, Recall, and F-Score on the intermediate set of Tanks \& Temples \cite{Knapitsch_2017_TNT}.}
\label{tab:tnt-prec-rec}
\end{table}

\begin{table*}[t]
\centering
\resizebox{0.85\textwidth}{!}{%
\begin{tabular}{l|c|ccccccc}
\hline
    \multirow{2}{*}{Method} &
    \multicolumn{7}{c}{BlendedMVS [MAE$\downarrow$]} \\ \cline{2-9}
    &
      \multicolumn{1}{c|}{Mean} &
      scan106 &
      scan107 &
      scan108 &
      scan109 &
      scan110 &
      scan111 &
      scan112 \\ \hline\hline
    GBi-Net \cite{Mi_2022_CVPR} &
      0.319 &
      1.661 &
      0.006 &
      0.027 &
      0.017 &
      0.025 &
      0.031 &
      0.462 \\
    + \network{} &
      \textbf{0.288} &
      \textbf{1.539} &
      \textbf{0.001} &
      \textbf{0.015} &
      \textbf{0.011} &
      \textbf{0.010} &
      \textbf{0.024} &
      \textbf{0.417} \\ \hline
\end{tabular}%
}
\caption{Quantitative comparison of the 2D depth map errors on the validation set of BlendedMVS \cite{yao2020blendedmvs} benchmark. The best results are marked in bold. Without any fine-tuning, \network{} improves the inputs of GBi-Net \cite{Mi_2022_CVPR} on all scenes in the validation set.}
\label{tab:blended-2d-error}
\end{table*}

\newcommand\wsddn{0.19\textwidth}
\begin{figure*}[t]
\footnotesize
    \centering
    \begin{tabular}{ccccc}
        \includegraphics[width=\wsddn]{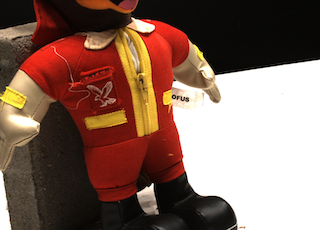} & \hspace{-10pt}
        \includegraphics[width=\wsddn]{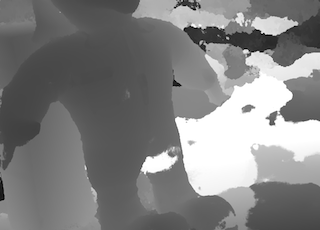} & \hspace{-10pt}
        \includegraphics[width=\wsddn]{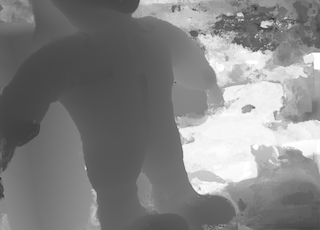} & \hspace{-10pt}
        \includegraphics[width=\wsddn]{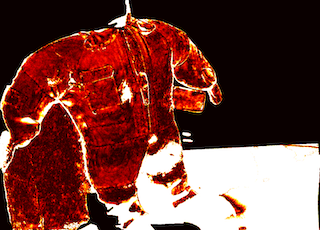} & \hspace{-10pt}
        \includegraphics[width=\wsddn]{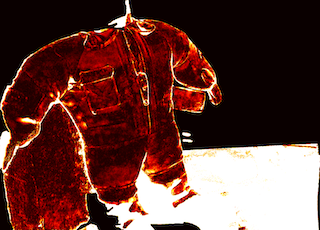} \\
        \includegraphics[width=\wsddn]{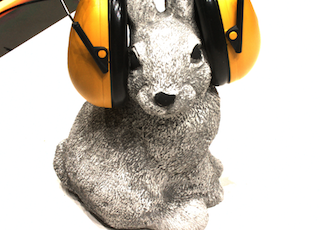} & \hspace{-10pt}
        \includegraphics[width=\wsddn]{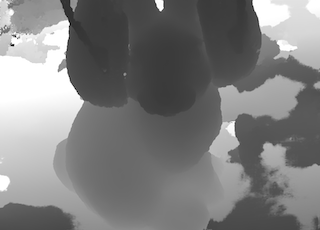} & \hspace{-10pt}
        \includegraphics[width=\wsddn]{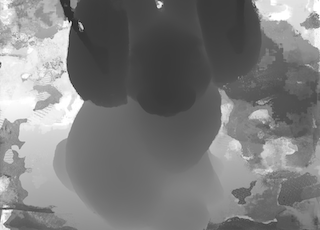} & \hspace{-10pt}
        \includegraphics[width=\wsddn]{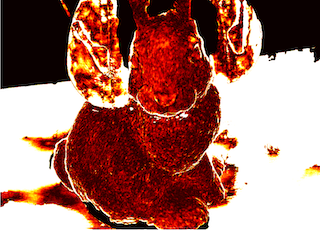} & \hspace{-10pt}
        \includegraphics[width=\wsddn]{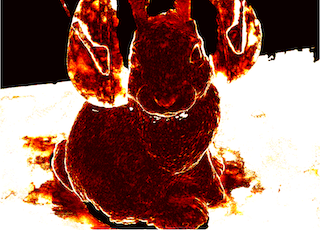} \\
        \includegraphics[width=\wsddn]{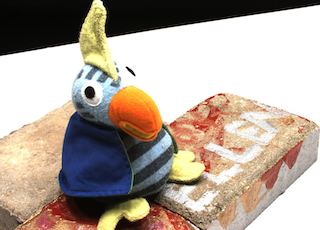} & \hspace{-10pt}
        \includegraphics[width=\wsddn]{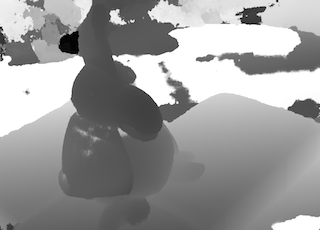} & \hspace{-10pt}
        \includegraphics[width=\wsddn]{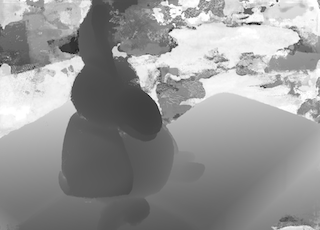} & \hspace{-10pt}
        \includegraphics[width=\wsddn]{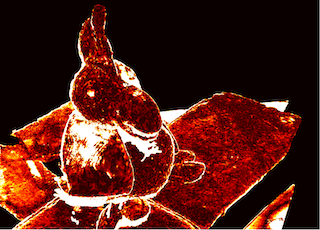} & \hspace{-10pt}
        \includegraphics[width=\wsddn]{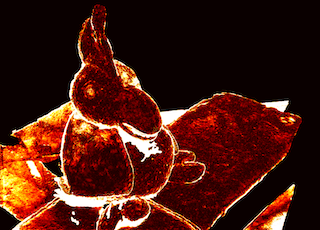} \\
        \includegraphics[width=\wsddn]{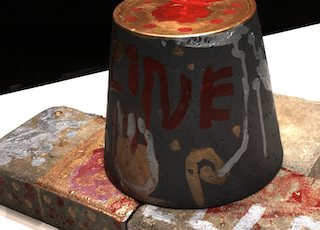} & \hspace{-10pt}
        \includegraphics[width=\wsddn]{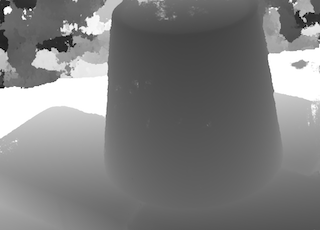} & \hspace{-10pt}
        \includegraphics[width=\wsddn]{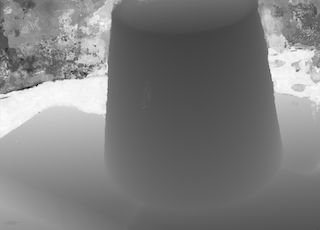} & \hspace{-10pt}
        \includegraphics[width=\wsddn]{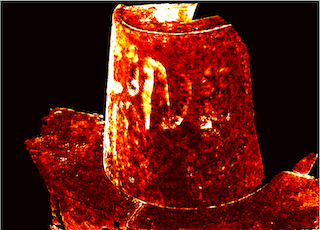} & \hspace{-10pt}
        \includegraphics[width=\wsddn]{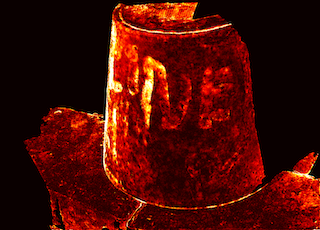} \\
        (a) Image & \hspace{-10pt}
        (b) NP-CVP-MVSNet Depth \cite{Yang_2022_CVPR} & \hspace{-10pt}
        (c) \network{} Depth & \hspace{-10pt}
        (d) NP-CVP-MVSNet Error \cite{Yang_2022_CVPR} & \hspace{-10pt}
        (e) \network{} Error
    \end{tabular}
    \caption{Qualitative comparison on the DTU \cite{aanaes16} dataset between NP-CVP-MVSNet \cite{Yang_2022_CVPR} and \network{}. We compare the input and fused depth and error maps. Error maps are colored using a heat map (larger errors correspond to brighter colors).}
    \label{fig:sup_dtu_depths_npcvp}
\end{figure*}

\newcommand\wcc{0.32\columnwidth}
\begin{figure}[t]
\footnotesize
    \centering
    \begin{tabular}{ccc}
        \includegraphics[width=\wcc]{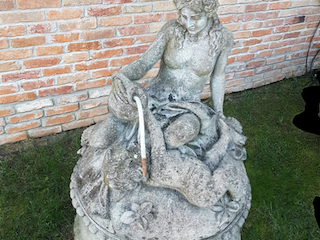} & \hspace{-10pt}
        \includegraphics[width=\wcc]{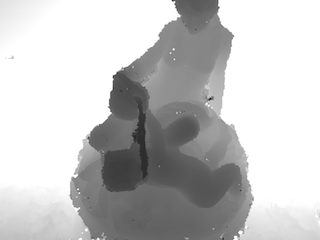} & \hspace{-10pt}
        \includegraphics[width=\wcc]{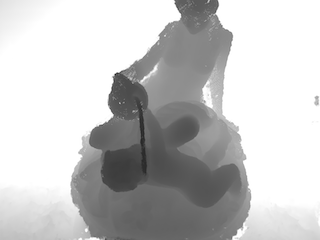} \\
        \includegraphics[width=\wcc]{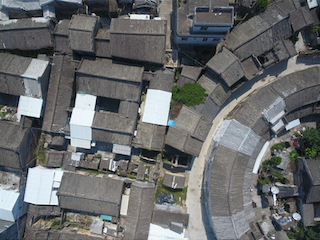} & \hspace{-10pt}
        \includegraphics[width=\wcc]{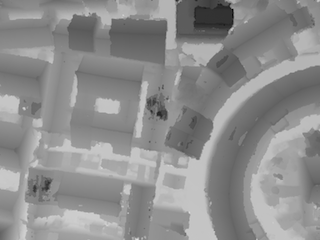} & \hspace{-10pt}
        \includegraphics[width=\wcc]{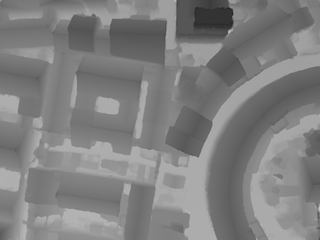} \\
        (c) Image & \hspace{-10pt}
        (d) Input Depth & \hspace{-10pt}
        (c) Fused Depth\\
    \end{tabular}
    \caption{Qualitative examples comparing the input depth maps with the fused output depth maps from the BlendedMVS \cite{yao2020blendedmvs} dataset using GBi-Net \cite{Mi_2022_CVPR} as input.}
    \label{fig:blended_qual}
\end{figure}

\newcommand\wnfc{0.24\textwidth}
\begin{figure*}[t]
\footnotesize
    \centering
    \begin{tabular}{cccc}
        \includegraphics[width=\wnfc]{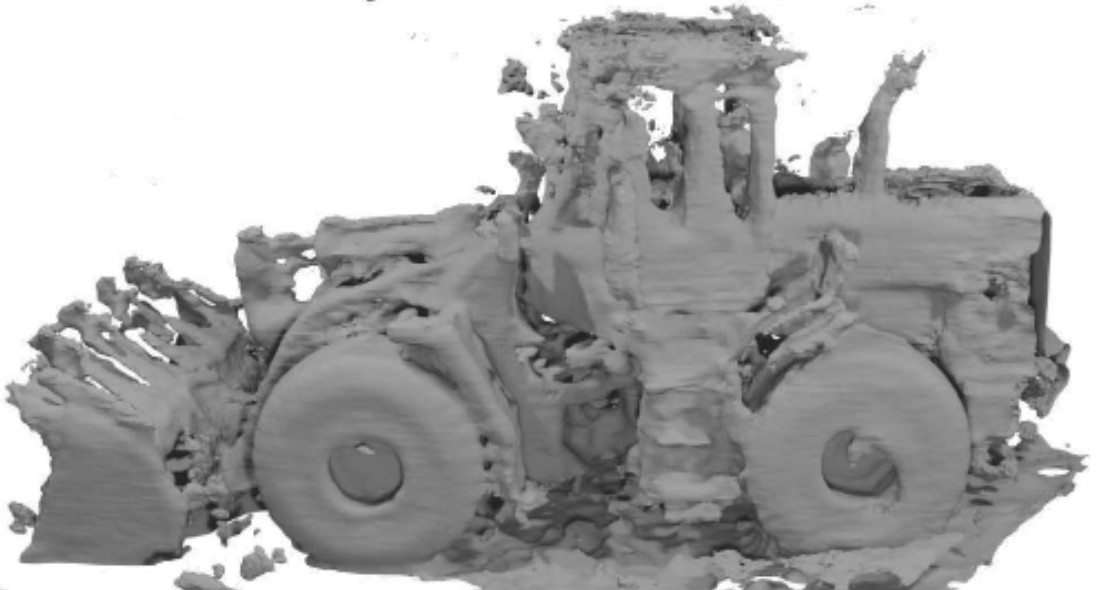} & 
        \includegraphics[width=\wnfc]{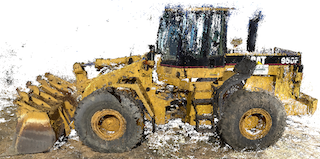} &
        \includegraphics[width=\wnfc]{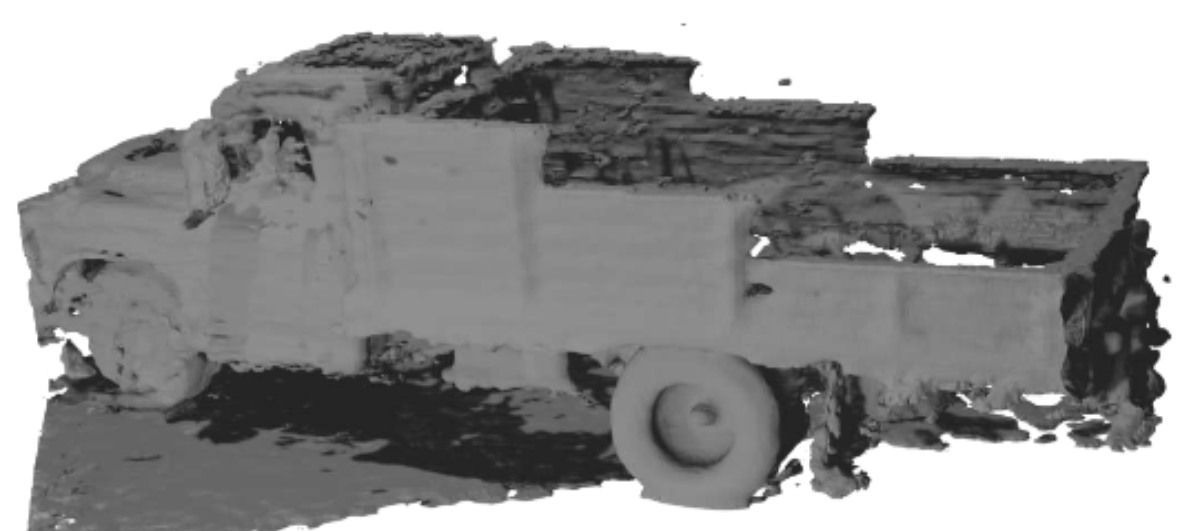} & 
        \includegraphics[width=\wnfc]{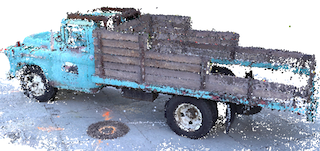} \\
        (a) NeuralFusion \cite{Weder_2021_CVPR} &
        (b) \network{} &
        (c) NeuralFusion \cite{Weder_2021_CVPR} &
        (d) \network{}
    \end{tabular}
    \caption{Qualitative examples comparing the reconstruction results of NeuralFusion \cite{Weder_2021_CVPR} and \network{} on scenes from Tanks \& Temples \cite{Knapitsch_2017_TNT}. The visuals of the 3D models produced by NeuralFusion \cite{Weder_2021_CVPR} are sampled from the paper.}
    \label{fig:neural_fusion_comp}
\end{figure*}

\section{Ablation Studies}
\label{sec:sup_ablations}
We provide ablation studies evaluating the contributions of several aspects of the network architecture. In Table \ref{tab:dtu-2d-ablations}, we evaluate the MAE isolating the SWE sub-network, as well as the different constraints. We show the contributions of using the brute-force search space approach, as well as the contributions of using only support and only occlusions and free-space violations. In Table \ref{tab:swe-ablations}, we show the memory, run-time, and parameter difference between the brute-force approach and the SWE sub-network.

Most \textit{state-of-the-art} Deep MVS architectures directly compute confidence estimates from the output probability volume of the network, using a small window around the estimated depth voxel. We explore using this method, incorporating the radius estimate from the SWE sub-network, as well as using only the radius to compute confidence. Specifically, following previous work in Deep MVS, we compute confidence maps from the output probability volume of the network by summing the probability values for the four surrounding voxels corresponding to the index of the selected depth. We then test adding the inverse of the radius value from the SWE sub-network as a weighting to this confidence. Finally, we test using only the inverse radius value to compute our confidence.  We evaluate these different methods of producing confidence maps in Table \ref{tab:auc-ablations}. In all instances, the confidence maps produced by \network{} are better indications of depth estimate quality, and can be used to more effectively rank depth estimates.

\section{Additional Quantitative Evaluations}
\label{sec:sup_quant}
We provide additional results on the \textbf{BlendedMVS} \cite{yao2020blendedmvs} dataset. We report the MAE on the validation set between the GBi-Net \cite{Mi_2022_CVPR} input depth maps and the \network{} output fused depth maps. We used the pre-trained GBi-Net model trained on the BlendedMVS training set and tested \network{} using the model trained on DTU without any fine-tuning. The quantitative results are presented in Table \ref{tab:blended-2d-error}. We show significant improvements in all scenes. Qualitative results can be viewed in Figure \ref{fig:blended_qual}.

We also provide the \textit{Precision} and \textit{Recall}, alongside the \textit{F-Score} on the \textbf{Tanks \& Temples} \cite{Knapitsch_2017_TNT} intermediate test set, retrieved from the benchmark leaderboard. The \textit{Precision} score is the percentage of points in the reconstructed point cloud that have a Chamfer distance to the closest point in the ground-truth point cloud below some threshold, $\tau$. The \textit{Recall} score is the percentage of points in the ground truth-point cloud that have a Chamfer distance to the closest point in the reconstructed point cloud below the same threshold, $\tau$. The \textit{F-Score} is then the harmonic mean of these two scores. Quantitative results can be found in Table \ref{tab:tnt-prec-rec}. We improve the Precision and F-Score using UCSNet as input, and improve the Recall using GBi-Net as input.

\section{Additional Qualitative Results}
\label{sec:sup_qual}
We show additional qualitative results for all baselines. Figure \ref{fig:sup_dtu_depths_npcvp} shows a comparison of depth and error maps between NP-CVP-MVSNet \cite{Yang_2022_CVPR} and \network{} on several scenes from the DTU \cite{aanaes16} evaluation set. In Figure \ref{fig:neural_fusion_comp}, we provide a comparison between the final 3D reconstruction results of NeuralFusion \cite{Weder_2021_CVPR} and \network{}. We only provide a qualitative comparison, as NeuralFusion does not provide any quantitative results on any of the datasets used in our experiments. Figure \ref{fig:sup_dtu_depths_gbinet} shows a comparison of depth and confidence maps between GBi-Net \cite{Mi_2022_CVPR} and \network{} on several scenes from the DTU evaluation set. In Figure \ref{fig:sup_tnt_depths}, we can observe the depth maps comparison of scenes from the Tanks \& Temples \cite{Knapitsch_2017_TNT} dataset between GBi-Net \cite{Mi_2022_CVPR} and \network{}. We provide depth map comparisons from several scenes of the validation set from BlendedMVS \cite{yao2020blendedmvs} using GBi-Net \cite{Mi_2022_CVPR} as input in Figure \ref{fig:sup_blended_depths}. We also show final reconstructions from the DTU and Tanks \& Temples datasets in Figure \ref{fig:sup_dtu_points} and Figure \ref{fig:sup_tnt_points}, respectively.

\newcommand\wsddg{0.19\textwidth}
\begin{figure*}[b]
\footnotesize
    \centering
    \begin{tabular}{ccccc}
        \includegraphics[width=\wsddg]{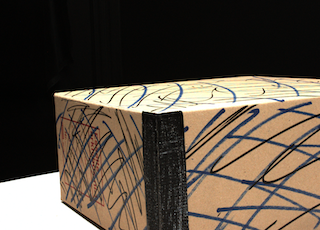} & \hspace{-10pt}
        \includegraphics[width=\wsddg]{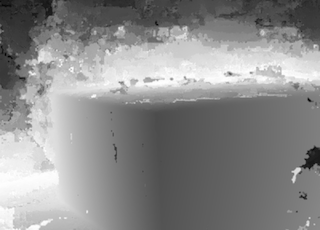} & \hspace{-10pt}
        \includegraphics[width=\wsddg]{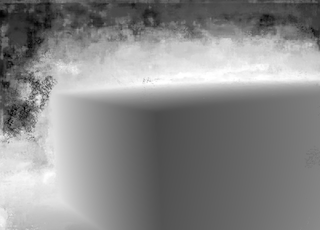} & \hspace{-10pt}
        \includegraphics[width=\wsddg]{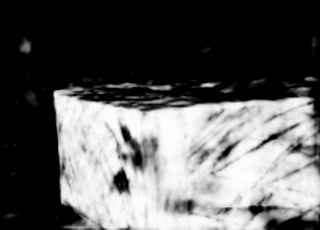} & \hspace{-10pt}
        \includegraphics[width=\wsddg]{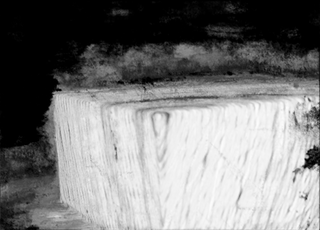} \\
        \includegraphics[width=\wsddg]{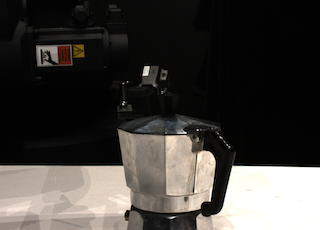} & \hspace{-10pt}
        \includegraphics[width=\wsddg]{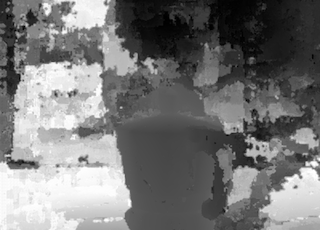} & \hspace{-10pt}
        \includegraphics[width=\wsddg]{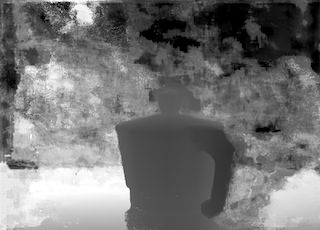} & \hspace{-10pt}
        \includegraphics[width=\wsddg]{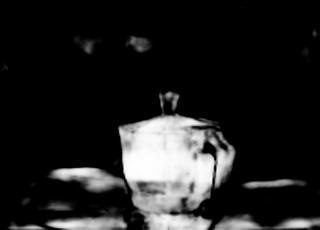} & \hspace{-10pt}
        \includegraphics[width=\wsddg]{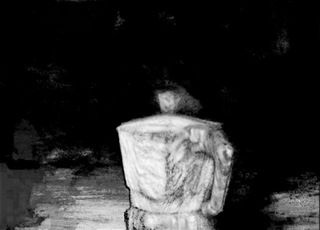} \\
        \includegraphics[width=\wsddg]{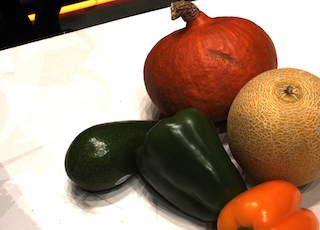} & \hspace{-10pt}
        \includegraphics[width=\wsddg]{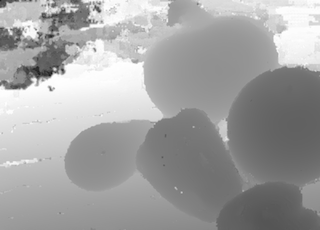} & \hspace{-10pt}
        \includegraphics[width=\wsddg]{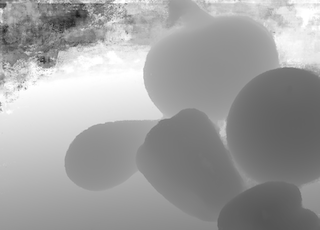} & \hspace{-10pt}
        \includegraphics[width=\wsddg]{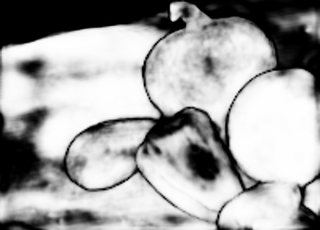} & \hspace{-10pt}
        \includegraphics[width=\wsddg]{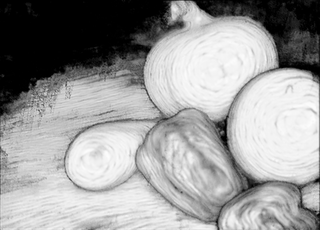} \\
        (a) Image & \hspace{-10pt}
        (b) GBi-Net Depth \cite{Mi_2022_CVPR} & \hspace{-10pt}
        (c) \network{} Depth & \hspace{-10pt}
        (d) GBi-Net Confidence \cite{Mi_2022_CVPR} & \hspace{-10pt}
        (e) \network{} Confidence
    \end{tabular}
    \caption{Qualitative comparison on the DTU \cite{aanaes16} dataset between GBi-Net \cite{Mi_2022_CVPR} and \network{}. We compare the input and fused depth and confidence maps. The improvements in surface boundary definition in the fused depth maps are also present in the fused confidence maps. We can observe much more detailed confidence maps, with more continuous changes in confidence values as opposed to very abrupt changes in the input confidence maps.}
    \label{fig:sup_dtu_depths_gbinet}
\end{figure*}

\newcommand\wstd{0.31\textwidth}
\begin{figure*}[b]
\footnotesize
    \centering
    \begin{tabular}{ccc}
        \includegraphics[width=\wstd]{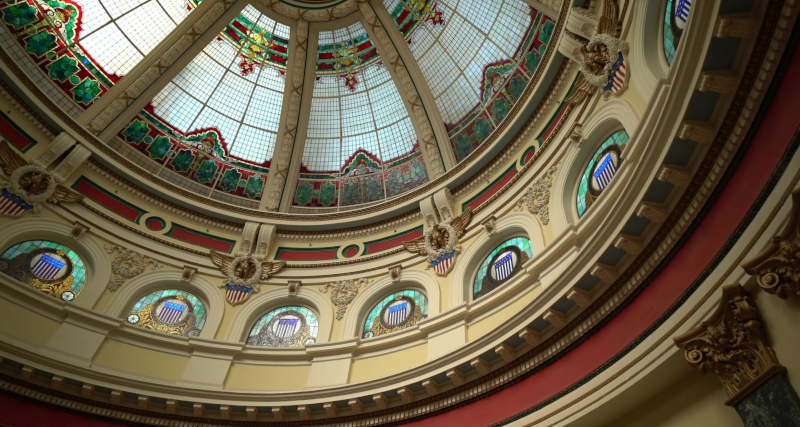} & \hspace{-10pt}
        \includegraphics[width=\wstd]{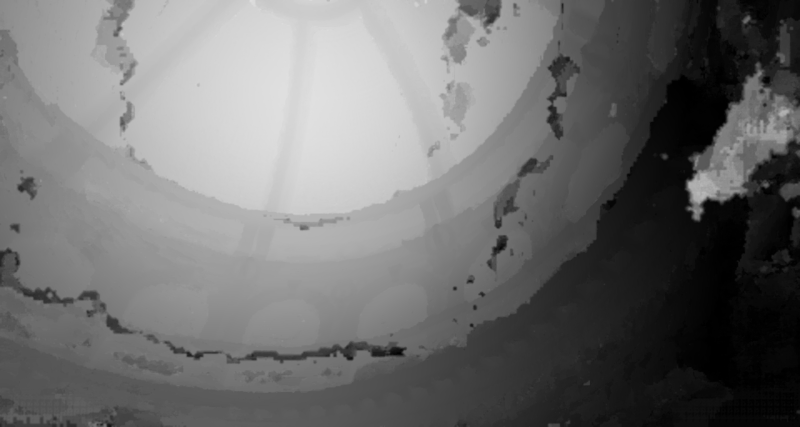} & \hspace{-10pt}
        \includegraphics[width=\wstd]{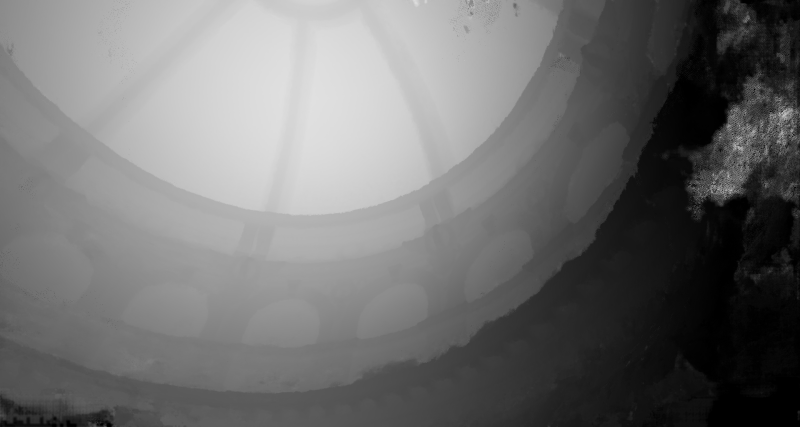} \\
        \includegraphics[width=\wstd]{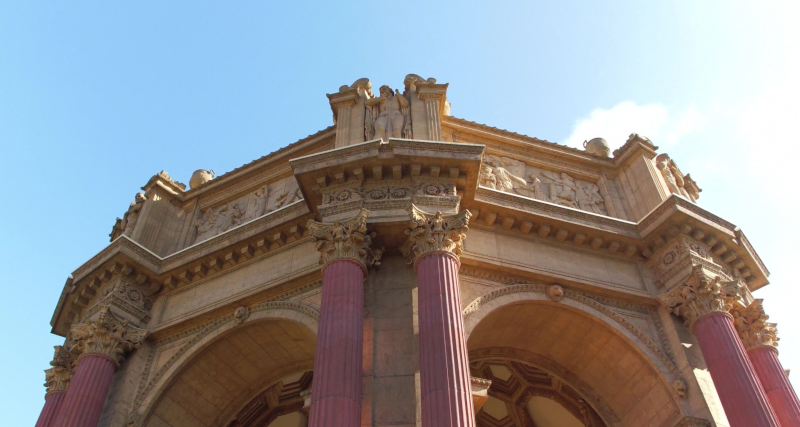} & \hspace{-10pt}
        \includegraphics[width=\wstd]{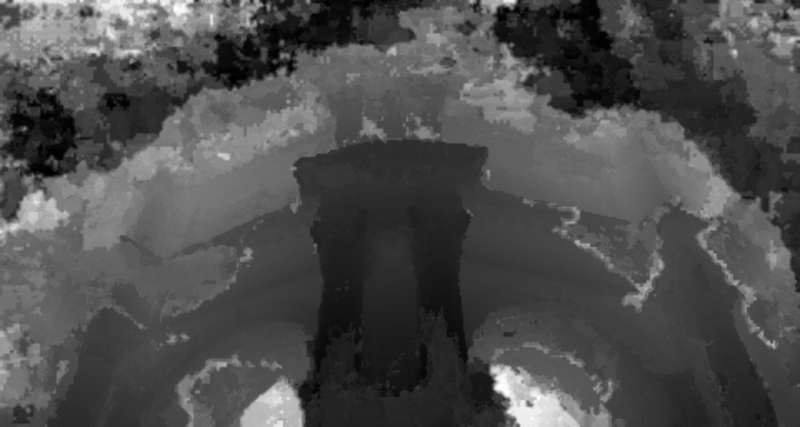} & \hspace{-10pt}
        \includegraphics[width=\wstd]{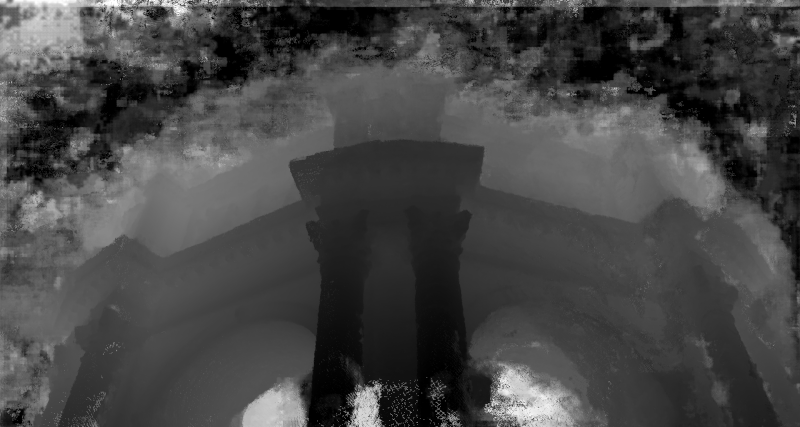} \\
        \includegraphics[width=\wstd]{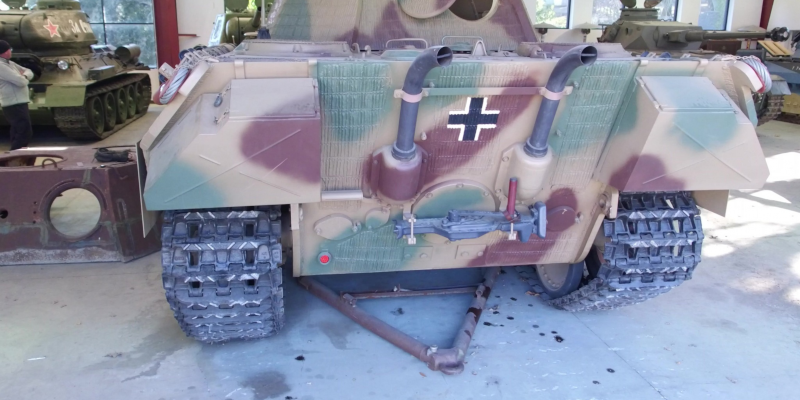} & \hspace{-10pt}
        \includegraphics[width=\wstd]{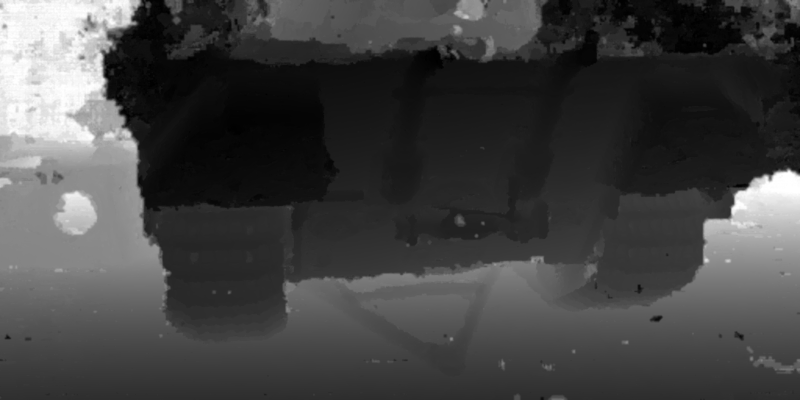} & \hspace{-10pt}
        \includegraphics[width=\wstd]{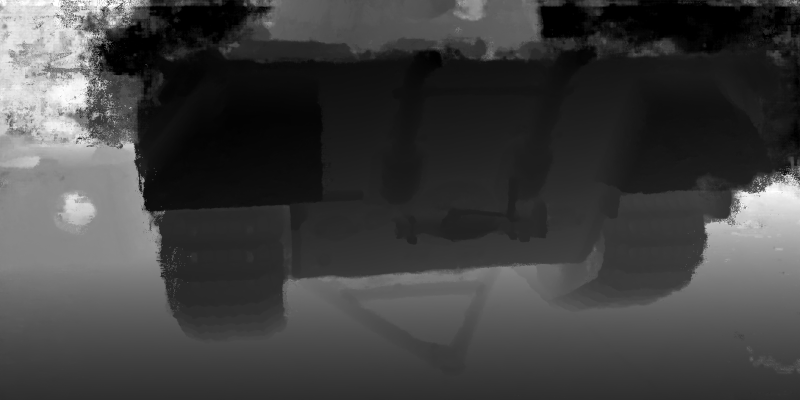} \\
        (a) Image & \hspace{-10pt}
        (b) GBi-Net \cite{Mi_2022_CVPR} Depth & \hspace{-10pt}
        (c) \network{} Depth
    \end{tabular}
    \caption{Qualitative depth map comparison on the Tanks \& Temples \cite{Knapitsch_2017_TNT} dataset between GBi-Net \cite{Mi_2022_CVPR} and \network{}.}
    \label{fig:sup_tnt_depths}
\end{figure*}

\newcommand\sswcc{0.32\textwidth}
\begin{figure*}[t]
\footnotesize
    \centering
    \begin{tabular}{ccc}
        \includegraphics[width=\sswcc]{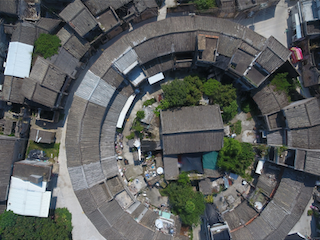} & \hspace{-10pt}
        \includegraphics[width=\sswcc]{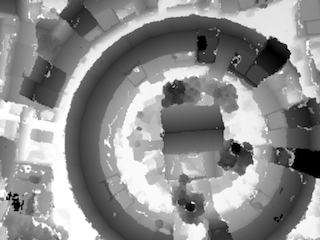} & \hspace{-10pt}
        \includegraphics[width=\sswcc]{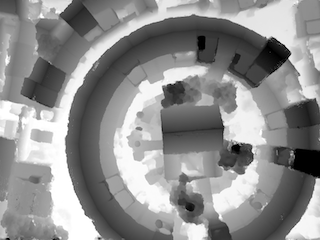} \\
        \includegraphics[width=\sswcc]{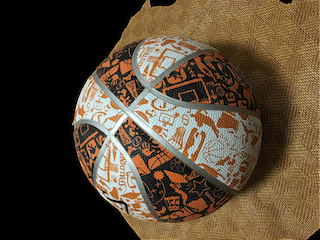} & \hspace{-10pt}
        \includegraphics[width=\sswcc]{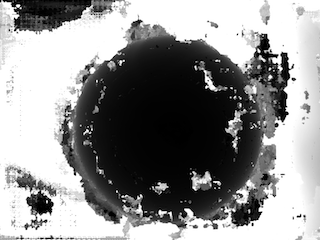} & \hspace{-10pt}
        \includegraphics[width=\sswcc]{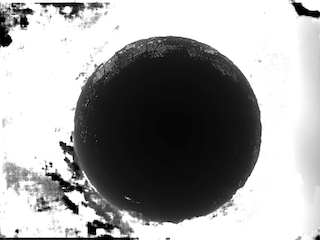} \\
        \includegraphics[width=\sswcc]{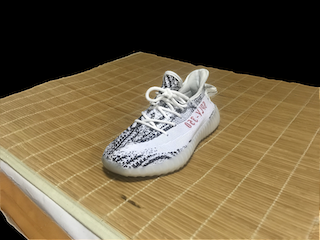} & \hspace{-10pt}
        \includegraphics[width=\sswcc]{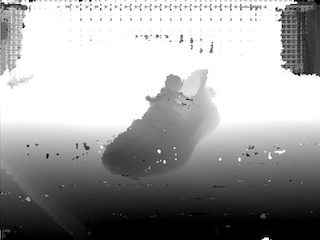} & \hspace{-10pt}
        \includegraphics[width=\sswcc]{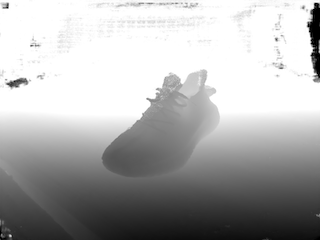} \\
        \includegraphics[width=\sswcc]{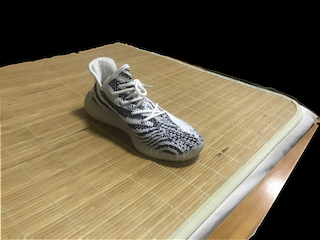} & \hspace{-10pt}
        \includegraphics[width=\sswcc]{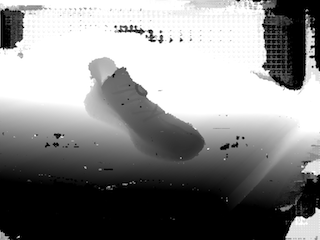} & \hspace{-10pt}
        \includegraphics[width=\sswcc]{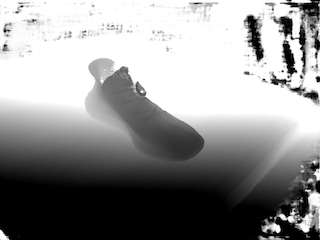} \\
        (c) Image & \hspace{-10pt}
        (d) GBi-Net \cite{Mi_2022_CVPR} Depth & \hspace{-10pt}
        (c) \network{} Depth\\
    \end{tabular}
    \caption{Qualitative examples comparing the input depth maps with the fused output depth maps from the BlendedMVS \cite{yao2020blendedmvs} dataset using GBi-Net \cite{Mi_2022_CVPR} as input.}
    \label{fig:sup_blended_depths}
\end{figure*}

\newcommand\wsdp{0.24\textwidth}
\begin{figure*}[t]
\footnotesize
    \centering
    \begin{tabular}{cccc}
        \includegraphics[width=\wsdp]{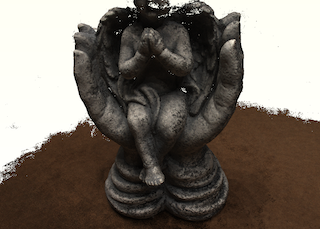} & \hspace{-10pt}
        \includegraphics[width=\wsdp]{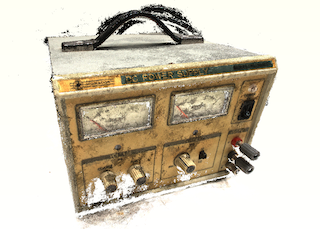} & \hspace{-10pt}
        \includegraphics[width=\wsdp]{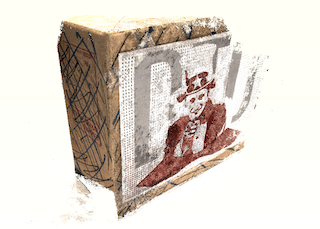} & \hspace{-10pt}
        \includegraphics[width=\wsdp]{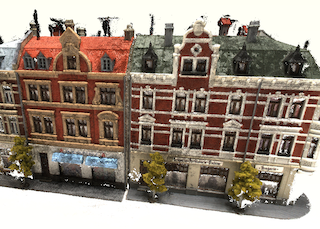} \\
        \includegraphics[width=\wsdp]{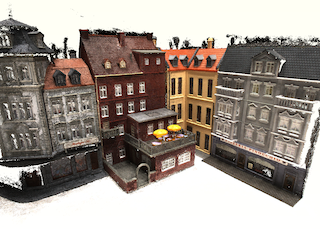} & \hspace{-10pt}
        \includegraphics[width=\wsdp]{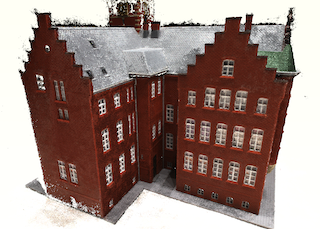} & \hspace{-10pt}
        \includegraphics[width=\wsdp]{assets/images/dtu/point_clouds/npcvp/scan29_00.png} & \hspace{-10pt}
        \includegraphics[width=\wsdp]{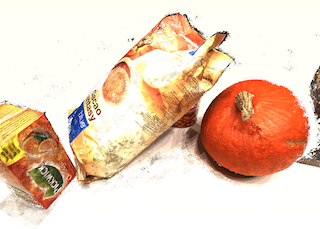} \\
        \includegraphics[width=\wsdp]{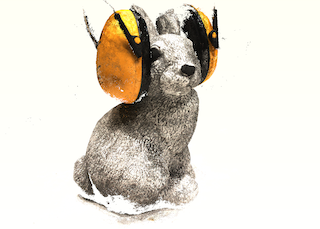} & \hspace{-10pt}
        \includegraphics[width=\wsdp]{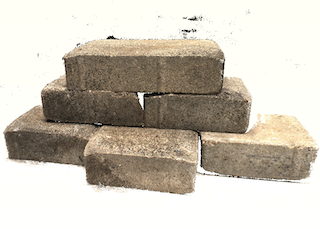} & \hspace{-10pt}
        \includegraphics[width=\wsdp]{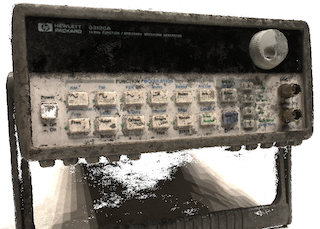} & \hspace{-10pt}
        \includegraphics[width=\wsdp]{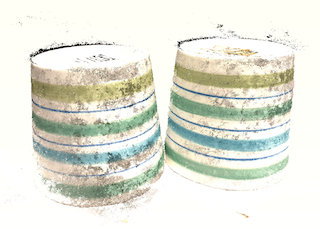} \\
        \includegraphics[width=\wsdp]{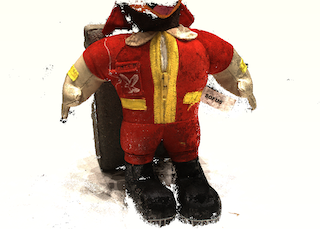} & \hspace{-10pt}
        \includegraphics[width=\wsdp]{assets/images/dtu/point_clouds/npcvp/scan62_00.png} & \hspace{-10pt}
        \includegraphics[width=\wsdp]{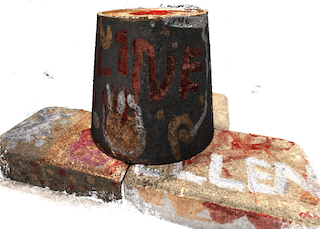} & \hspace{-10pt}
        \includegraphics[width=\wsdp]{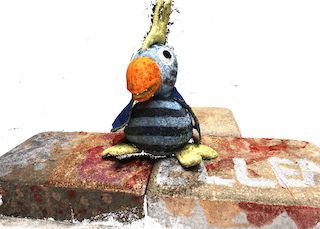} \\
        \includegraphics[width=\wsdp]{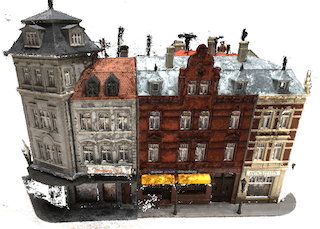} & \hspace{-10pt}
        \includegraphics[width=\wsdp]{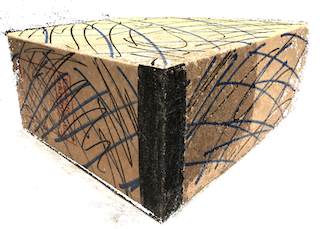} & \hspace{-10pt}
        \includegraphics[width=\wsdp]{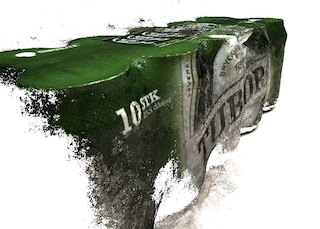} & \hspace{-10pt}
        \includegraphics[width=\wsdp]{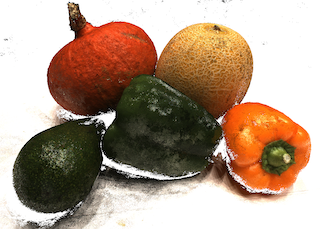} \\
        \includegraphics[width=\wsdp]{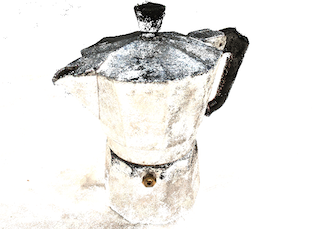} & \hspace{-10pt}
        \includegraphics[width=\wsdp]{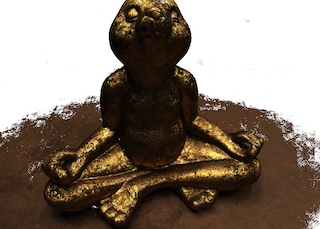} & \hspace{-10pt}
        \includegraphics[width=\wsdp]{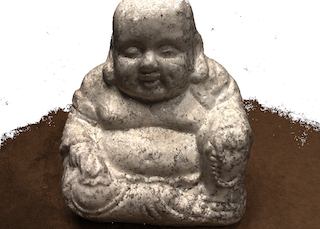} & \hspace{-10pt}
        \includegraphics[width=\wsdp]{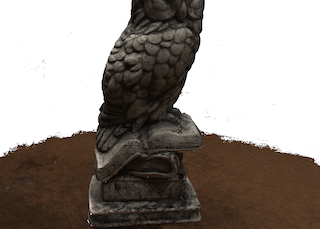} \\
    \end{tabular}
    \caption{Output point clouds of \network{} from the DTU \cite{aanaes16} dataset using NP-CVP-MVSNet \cite{Yang_2022_CVPR} as input.}
    \label{fig:sup_dtu_points}
\end{figure*}

\newcommand\wstp{0.32\textwidth}
\begin{figure*}[t]
\footnotesize
    \centering
    \begin{tabular}{ccccc}
        \includegraphics[width=\wstp]{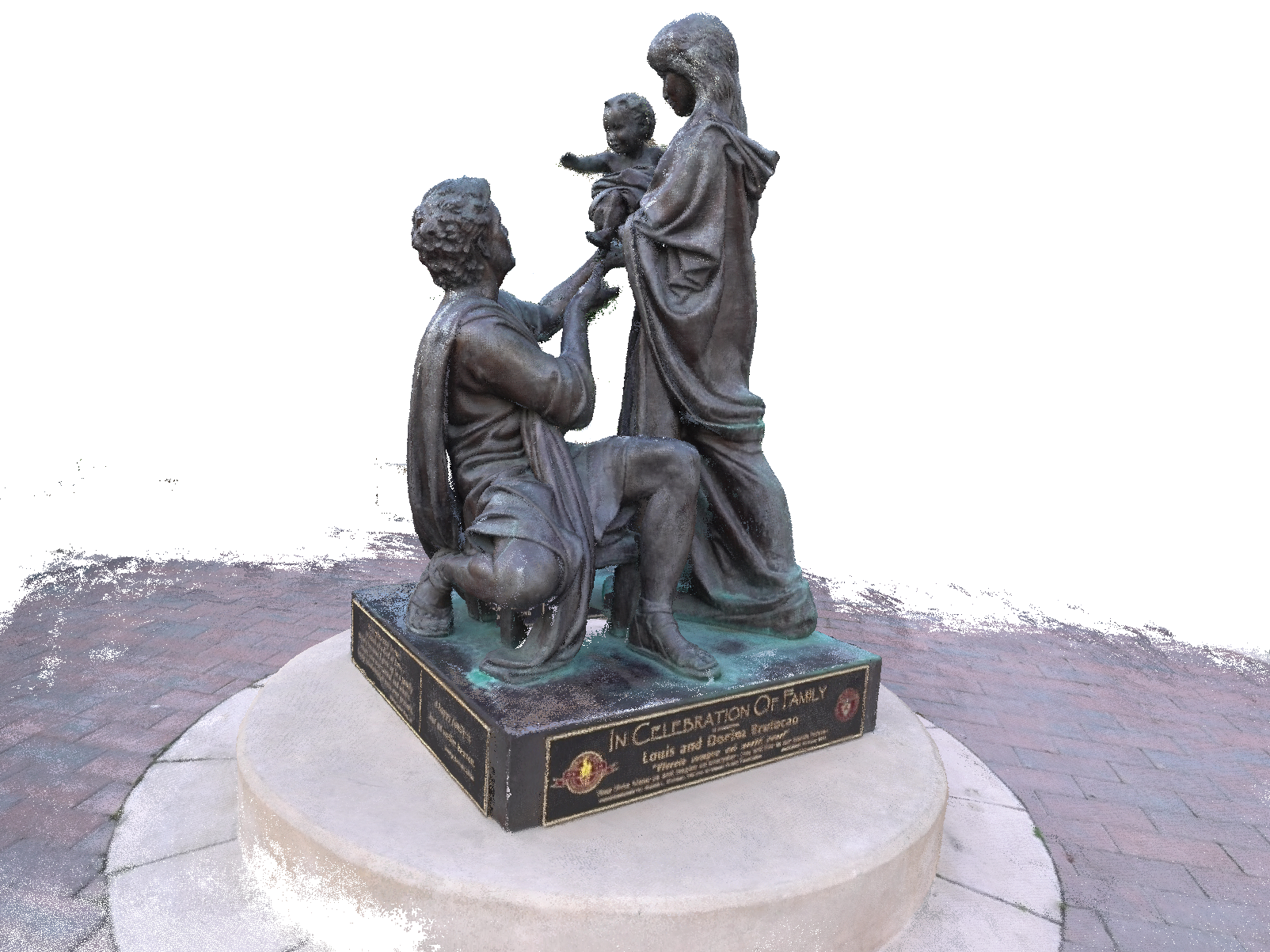} & \hspace{-10pt}
        \includegraphics[width=\wstp]{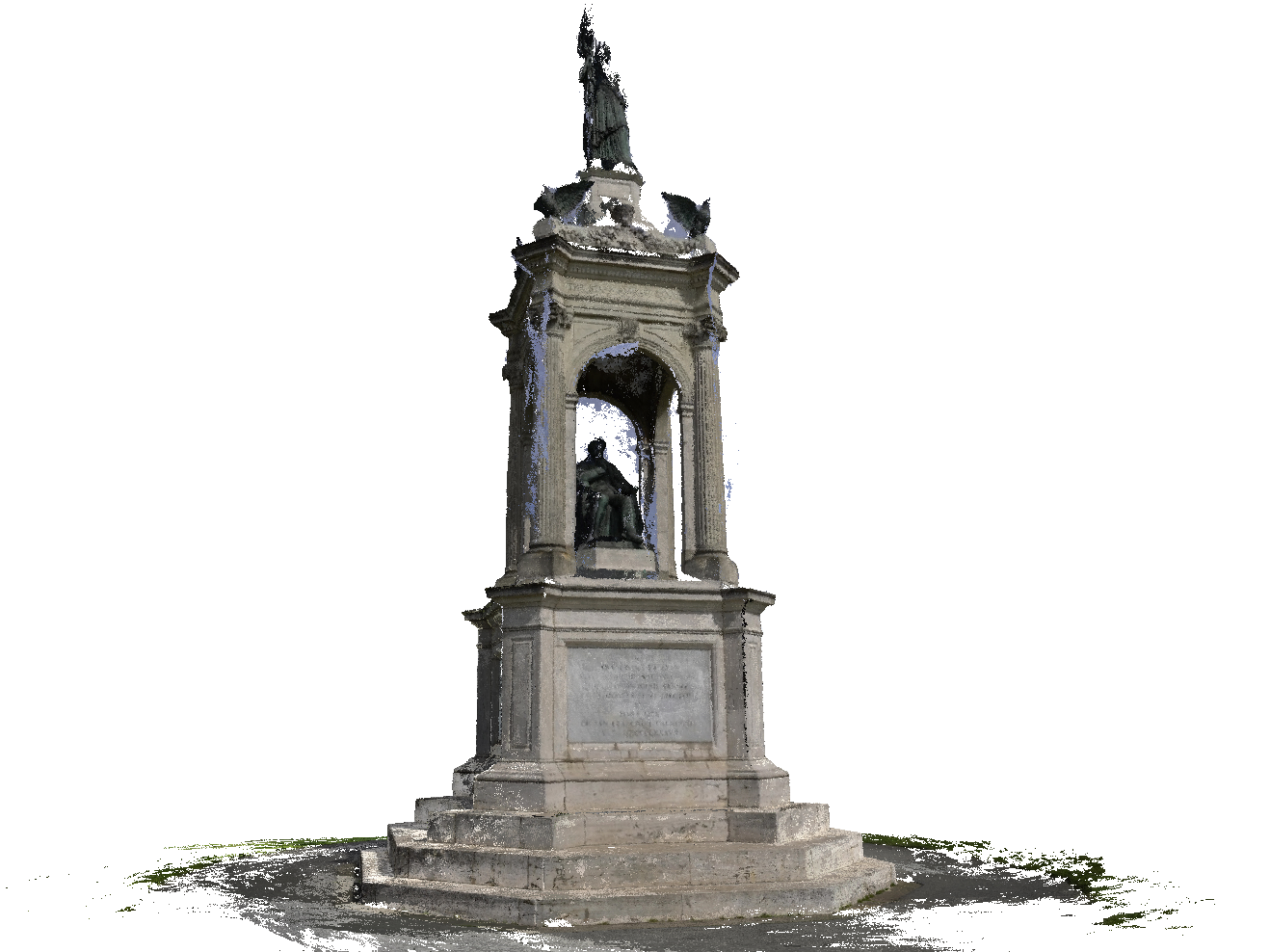} & \hspace{-10pt}
        \includegraphics[width=\wstp]{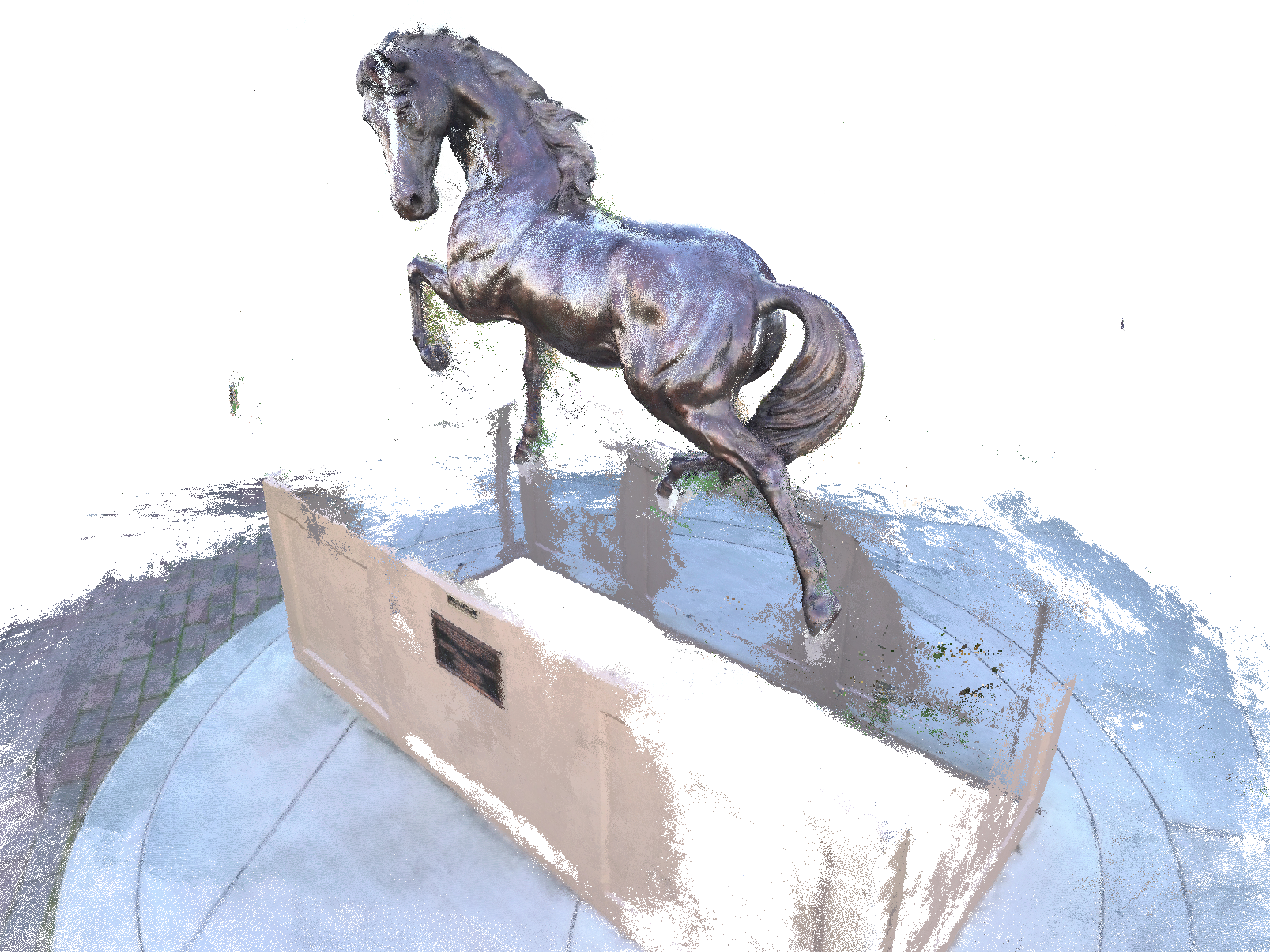} \\
        \includegraphics[width=\wstp]{assets/images/tnt/ucsnet/Family_01.png} & \hspace{-10pt}
        \includegraphics[width=\wstp]{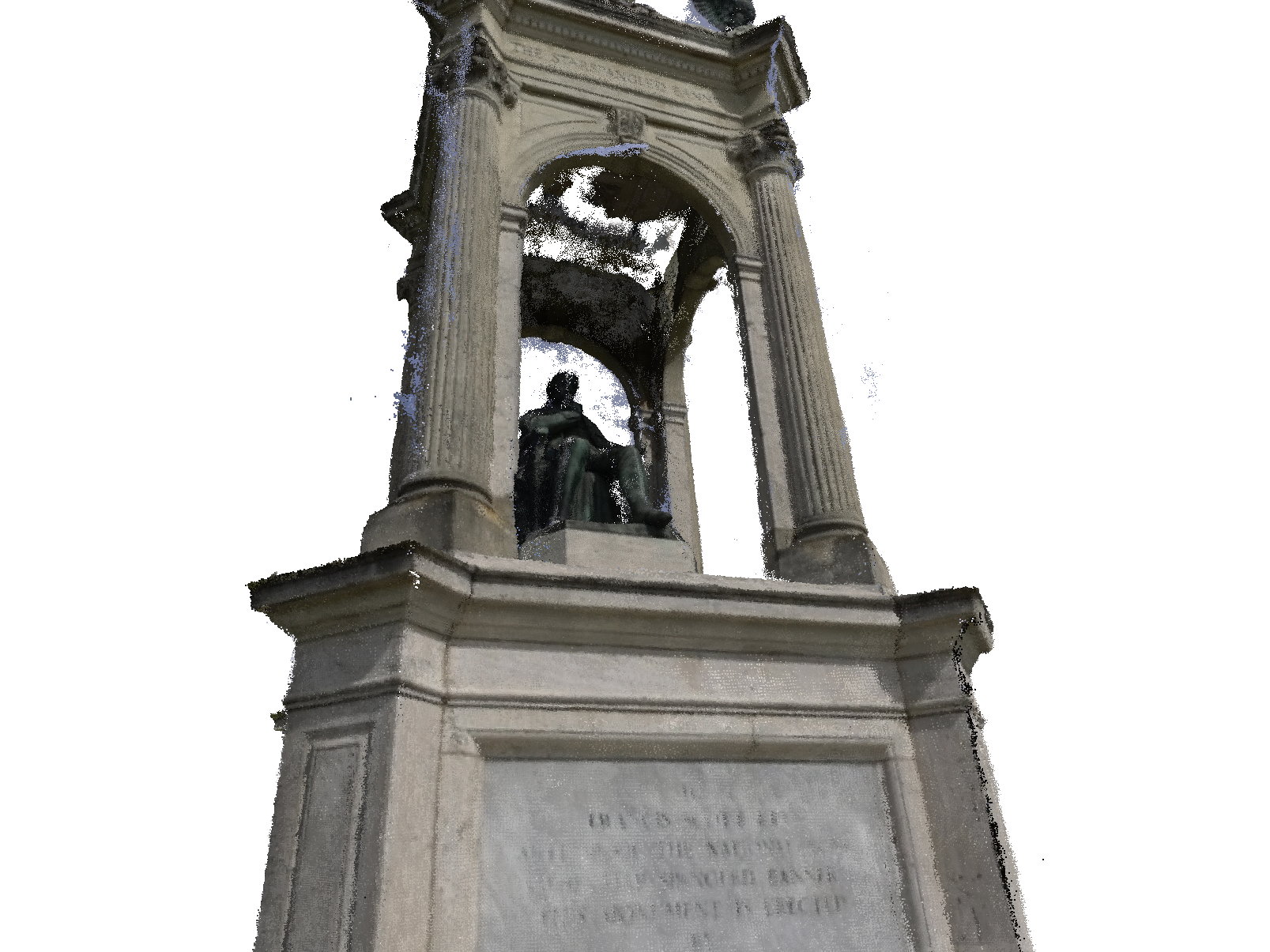} & \hspace{-10pt}
        \includegraphics[width=\wstp]{assets/images/tnt/ucsnet/Horse_01.png} \\
        \includegraphics[width=\wstp]{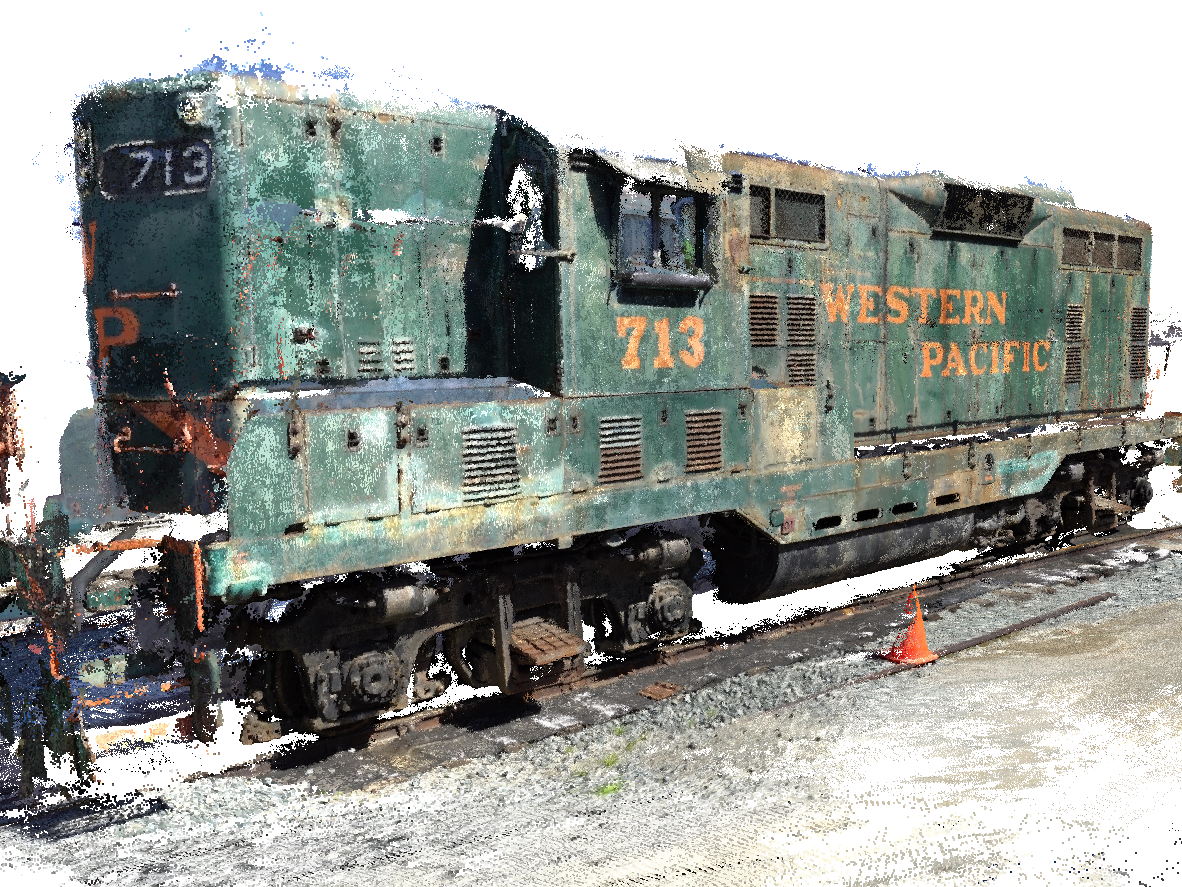} & \hspace{-10pt}
        \includegraphics[width=\wstp]{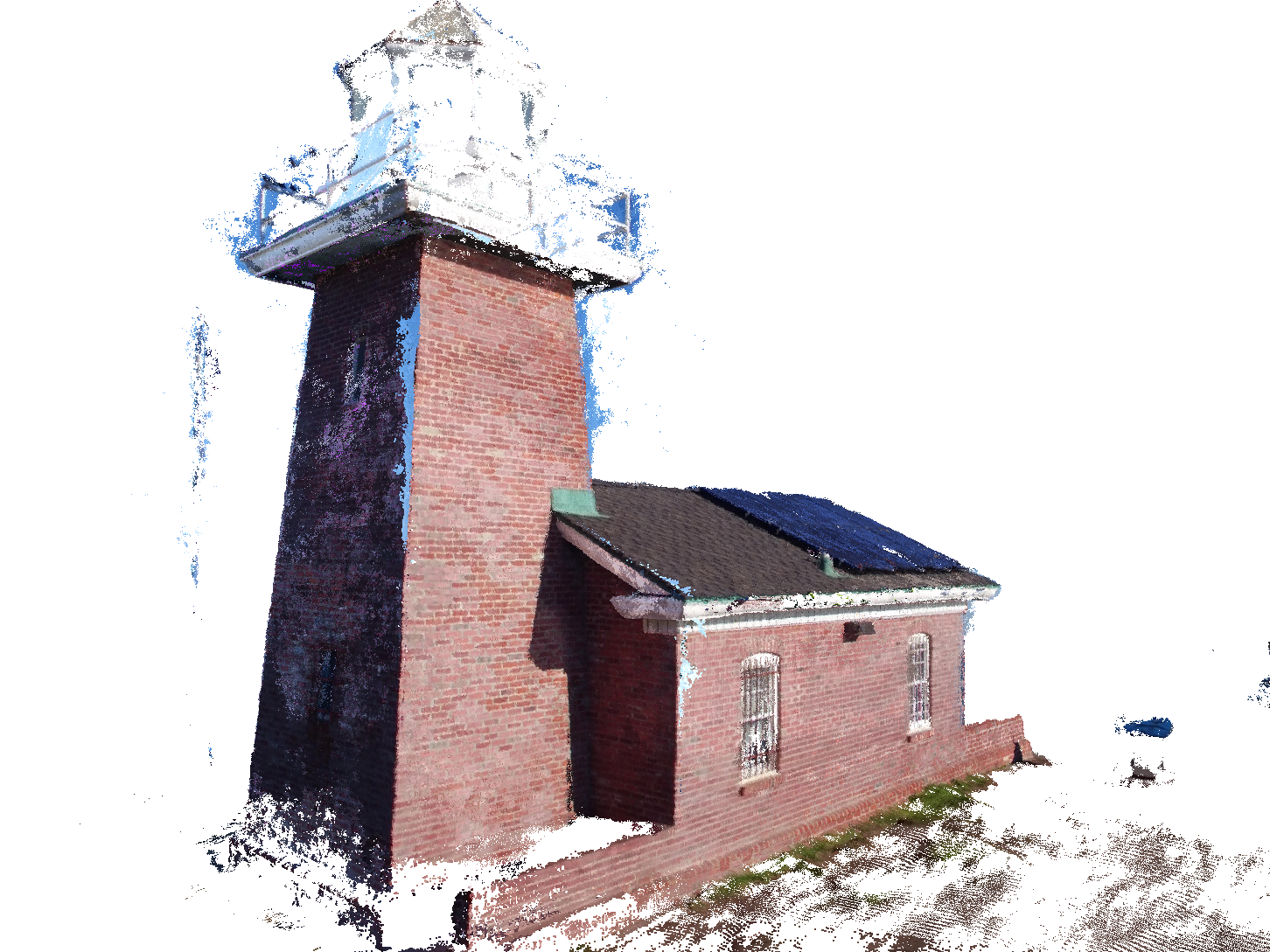} & \hspace{-10pt}
        \includegraphics[width=\wstp]{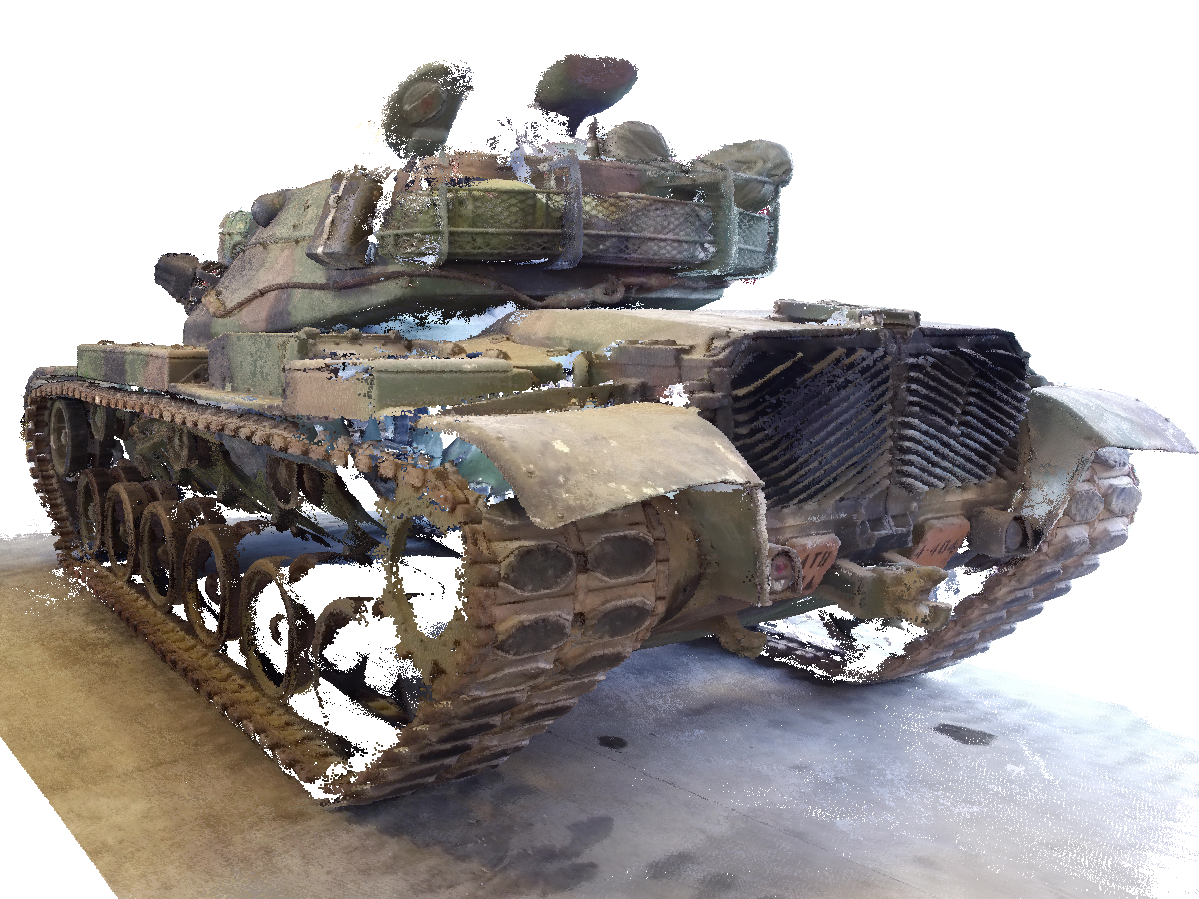} \\
        \includegraphics[width=\wstp]{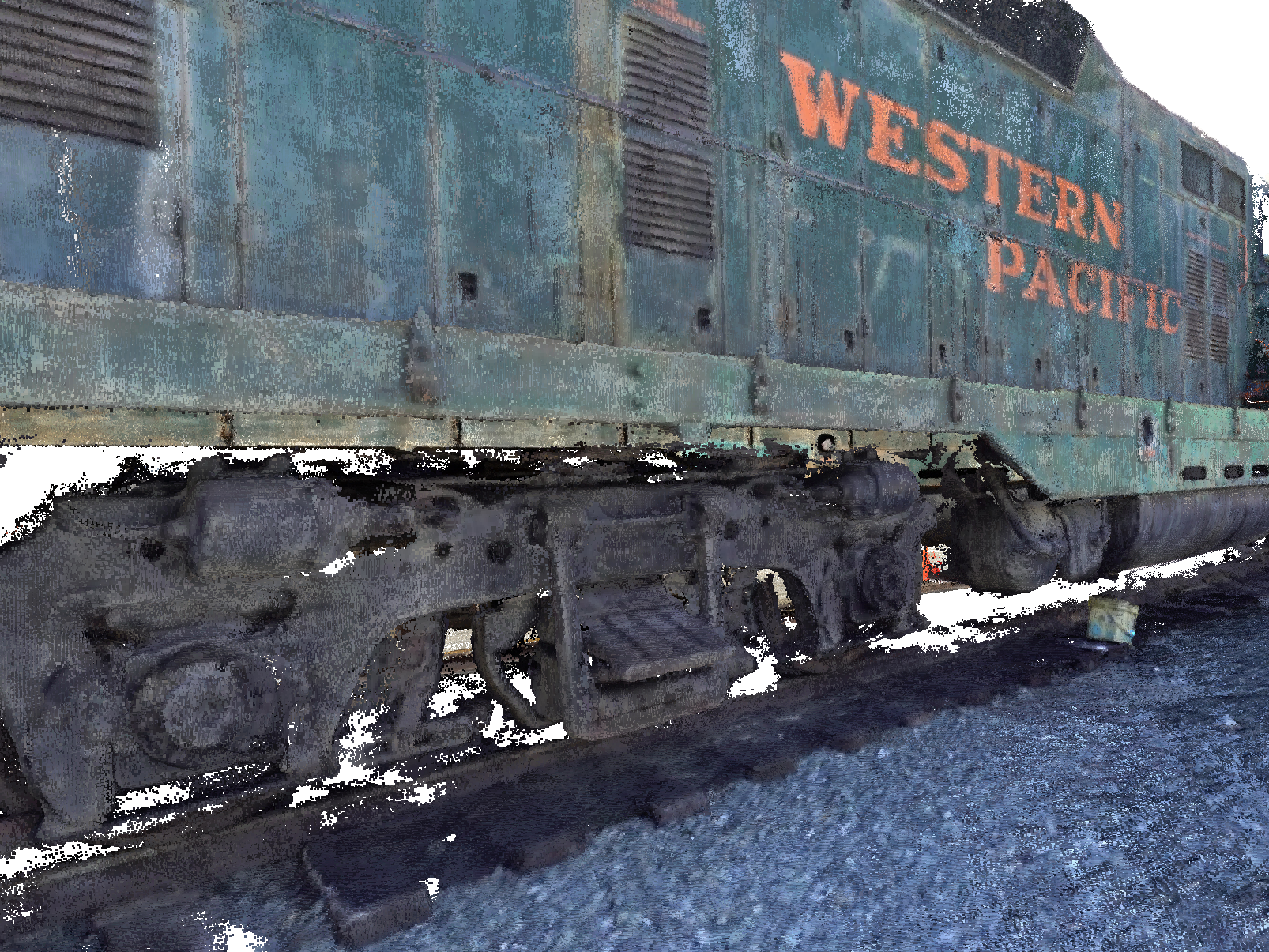} & \hspace{-10pt}
        \includegraphics[width=\wstp]{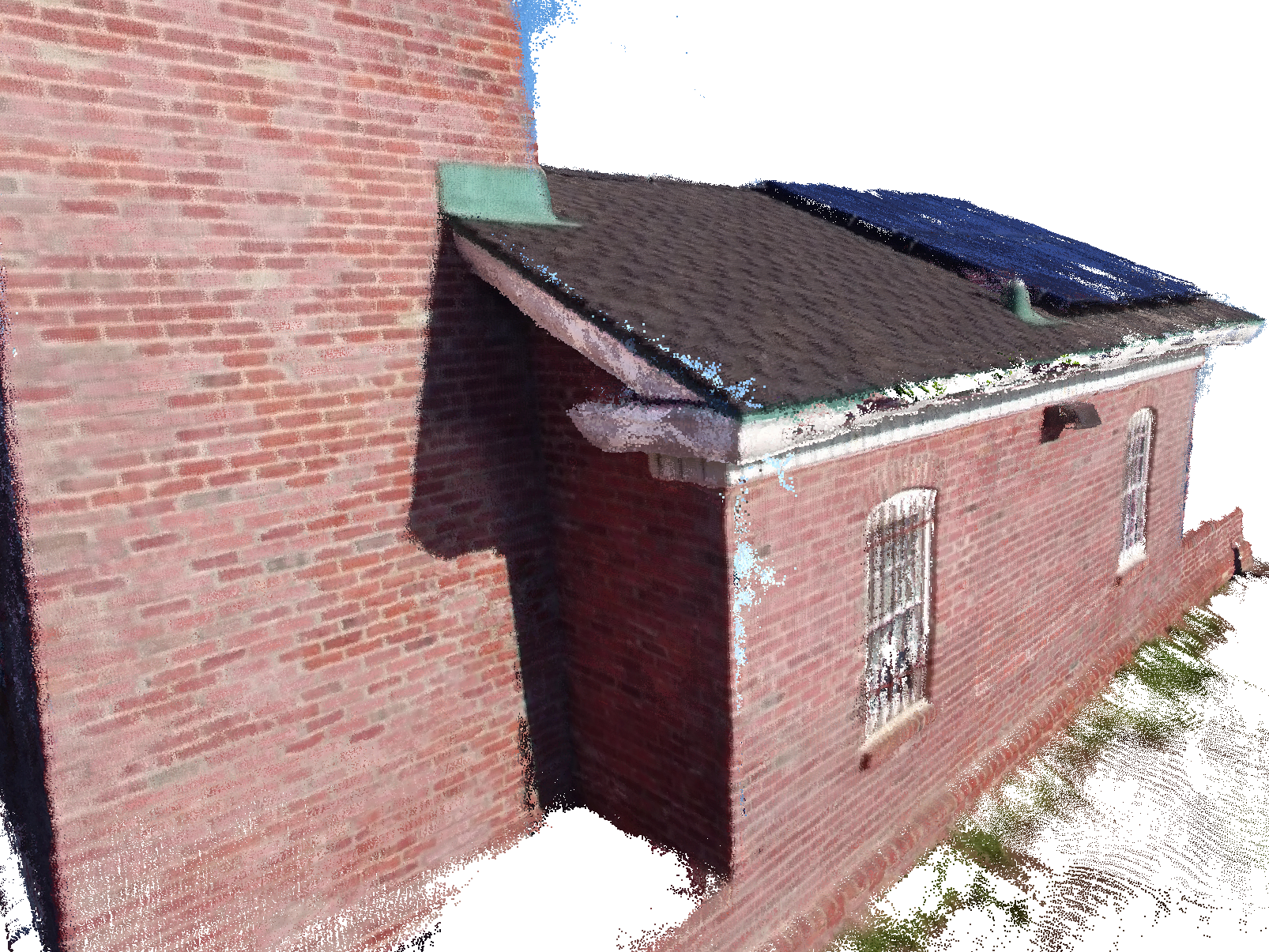} & \hspace{-10pt}
        \includegraphics[width=\wstp]{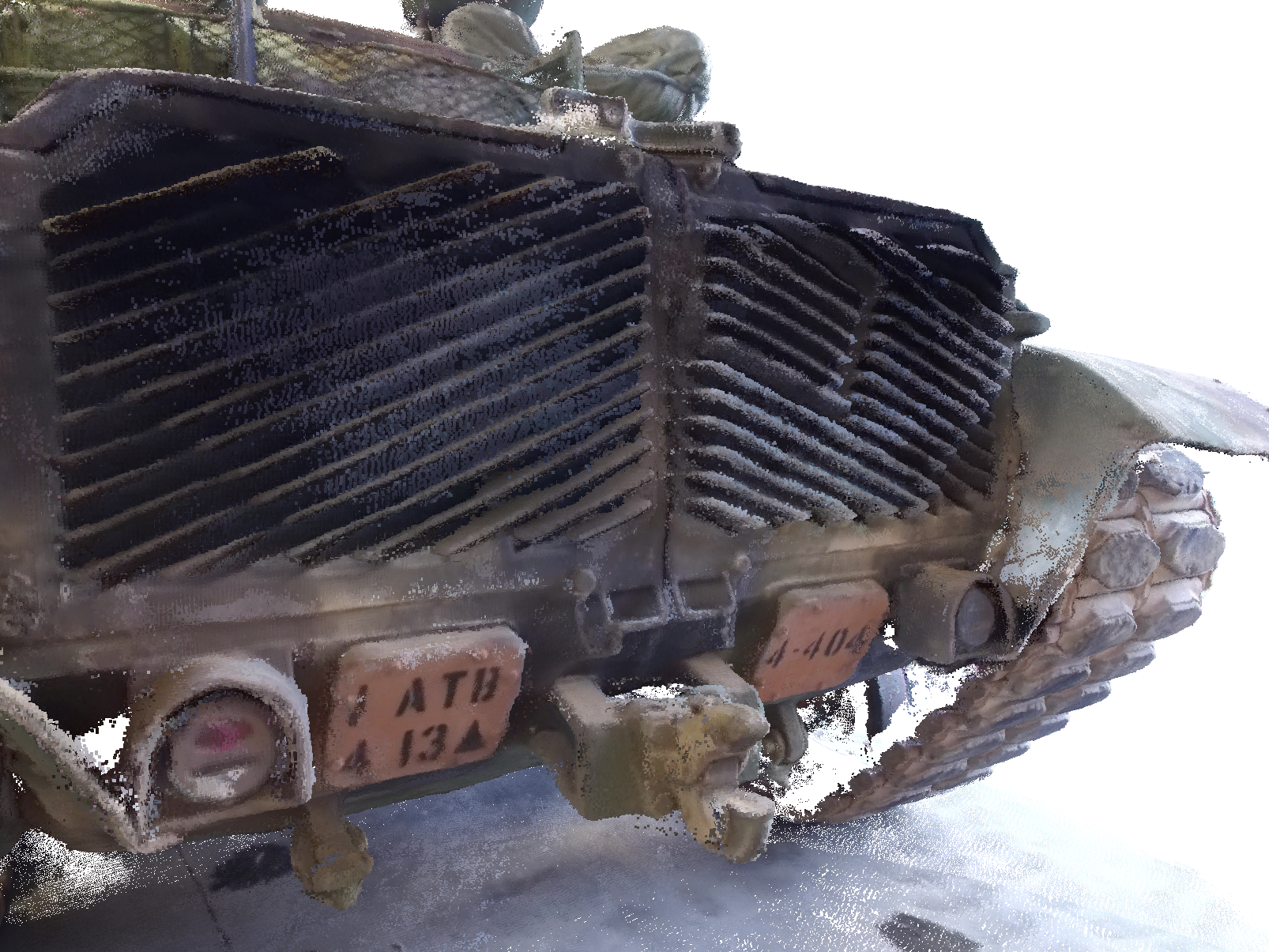} \\
    \end{tabular}
    \caption{Output point clouds of \network{} on the intermediate set from the Tanks \& Temples \cite{Knapitsch_2017_TNT} dataset using UCSNet \cite{Cheng_2020_CVPR} as input.}
    \label{fig:sup_tnt_points}
\end{figure*}